\documentclass[11pt]{article}
\usepackage{acl}
\usepackage{times}
\usepackage{latexsym}
\usepackage[T1]{fontenc}
\usepackage[utf8]{inputenc}
\usepackage{microtype}
\usepackage{inconsolata}
\usepackage{graphicx}
\usepackage[skins]{tcolorbox}
\usepackage{subcaption}
\usepackage{booktabs}
\usepackage{multirow}
\usepackage{makecell}
\usepackage{array}
\usepackage[english]{babel}
\usepackage{amsmath}
\usepackage{amssymb}
\usepackage{pifont}
\usepackage{xcolor}
\usepackage[normalem]{ulem}
\usepackage{flafter}
\usepackage{tikz}
\usetikzlibrary{shapes, backgrounds}
\newcolumntype{L}[1]{>{\raggedright\arraybackslash}p{#1}}
\newcommand{\squishlist}{
    \begin{list}{$\bullet$}
    { \setlength{\itemsep}{0pt}
        \setlength{\parsep}{1pt}
        \setlength{\topsep}{1pt}
        \setlength{\partopsep}{0pt}
        \setlength{\leftmargin}{1em}
        \setlength{\labelwidth}{1em}
        \setlength{\labelsep}{0.5em} } }
\newcommand{\squishend}{
    \end{list}  }
\newtcolorbox{insightbox}{
  enhanced,
  colback=teal!20,
  colframe=teal!90!black,
  boxrule=0pt,
  leftrule=5pt,
  rightrule=0pt,
  toprule=0pt,
  bottomrule=0pt,
  sharp corners,
  fontupper=\normalsize\raggedright,
  before skip=7pt,
  after skip=7pt,
  top=2pt,
  bottom=2pt,
}

\tcbset{
  appendixcard/.style={
    enhanced,
    coltext=black,
    fontupper=\small,
    boxrule=0.5mm,
    arc=3mm,
    top=1mm,
    bottom=1mm,
    left=1mm,
    right=1mm,
    boxsep=1.2mm
  }
}

\newcommand{\tabstrut}{\rule{0pt}{2.8ex}}
\definecolor{targetcolor}{HTML}{1565C0}
\definecolor{contrastcolor}{HTML}{E65100}
\newcommand{\tgt}[1]{\textcolor{targetcolor}{#1}}
\newcommand{\ctr}[1]{\textcolor{contrastcolor}{#1}}
\definecolor{metaURL}{HTML}{4E79A7}
\definecolor{metaCountry}{HTML}{2E7D32}
\definecolor{metaContinent}{HTML}{D97706}
\newcommand{\metachip}[3]{\begingroup\setlength{\fboxsep}{1.2pt}\colorbox{#1!10}{\textcolor{#1!68!black}{\scriptsize #2\,\textsf{#3}}}\endgroup}
\newcommand{\URLChip}{\metachip{metaURL}{\ding{108}}{URL}}
\newcommand{\CountryChip}{\metachip{metaCountry}{\ding{108}}{Country}}
\newcommand{\ContinentChip}{\metachip{metaContinent}{\ding{108}}{Continent}}
\newcommand{\URLCountryChip}{\URLChip{}\hspace{0.6pt}\CountryChip{}}
\newcommand{\URLContinentChip}{\URLChip{}\hspace{0.6pt}\ContinentChip{}}
\newcommand{\FullMetaChip}{\URLChip{}\hspace{0.6pt}\CountryChip{}\hspace{0.6pt}\ContinentChip{}}
\newcommand{\tableprimaryplus}[1]{\colorbox{globalwithtestwith!45}{#1}}
\newcommand{\tableprimaryminus}[1]{\colorbox{globalwithouttestwithout!22}{#1}}

\definecolor{cigainposstrongcolor}{HTML}{006D2C}
\definecolor{cigainposweakcolor}{HTML}{72AD75}
\definecolor{cigainnegstrongcolor}{HTML}{A50F15}
\definecolor{cigainnegweakcolor}{HTML}{D98787}
\definecolor{cigainzerocolor}{HTML}{A0A0A0}
\definecolor{cigaincicolor}{HTML}{B8B8B8}
\newcommand{\cigainci}[2]{\textcolor{cigaincicolor}{\tiny [#1,#2]}}
\newcommand{\cigainposstrong}[3]{\makecell[r]{\textcolor{cigainposstrongcolor}{\bfseries\boldmath #1}\\[-2.2pt]\cigainci{#2}{#3}}}
\newcommand{\cigainpos}[3]{\makecell[r]{\textcolor{cigainposweakcolor}{#1}\\[-2.2pt]\cigainci{#2}{#3}}}
\newcommand{\cigainnegstrong}[3]{\makecell[r]{\textcolor{cigainnegstrongcolor}{\bfseries\boldmath #1}\\[-2.2pt]\cigainci{#2}{#3}}}
\newcommand{\cigainneg}[3]{\makecell[r]{\textcolor{cigainnegweakcolor}{#1}\\[-2.2pt]\cigainci{#2}{#3}}}
\newcommand{\cigainzero}[3]{\makecell[r]{\textcolor{cigainzerocolor}{#1}\\[-2.2pt]\cigainci{#2}{#3}}}

\definecolor{mainbg}{HTML}{e7f5ff}
\definecolor{titlebg}{HTML}{a5d8ff}
\definecolor{fontcolor}{HTML}{1971c2}

\title{MAPLE: Metadata Conditioned LLM Pretraining\\for Locale-Aware Question Answering}
\author{
  Anjishnu Mukherjee \
  \textbf{Ziwei Zhu} \
  \textbf{Antonios Anastasopoulos} \\
  George Mason University\\
  \texttt{\{amukher6,zzhu20,antonis\}@gmu.edu}
}

\DeclareRobustCommand{\highlightRounded}[2]{%
  \begin{tikzpicture}[baseline=(word.base)]
    \node[
      rectangle,
      rounded corners,
      fill=#1,
      inner sep=2pt
    ] (word) {#2};
  \end{tikzpicture}%
}

\DeclareRobustCommand{\highlightRoundedBorder}[2]{%
  \begin{tikzpicture}[baseline=(word.base)]
    \node[
      rectangle,
      rounded corners,
      fill=#1,
      draw=black,
      line width=0.5pt,
      inner sep=2pt
    ] (word) {#2};
  \end{tikzpicture}%
}

\newcommand{\inlineicon}[2][0.95em]{%
  \raisebox{-0.18\height}{\includegraphics[height=#1]{#2}}%
}
\newcommand{\providericonfile}[1]{\inlineicon[0.95em]{figs/icons/providers/#1.png}}
\newcommand{\topiciconfile}[1]{\inlineicon[0.92em]{figs/icons/topics/#1.png}}
\newcommand{\flagiconfile}[1]{\inlineicon[0.84em]{figs/icons/flags/#1.png}}

\newcommand{\providerMeta}{\providericonfile{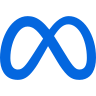}}
\newcommand{\providerQwen}{\providericonfile{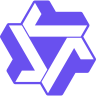}}
\newcommand{\providerGoogle}{\providericonfile{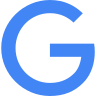}}

\newcommand{\providerMistral}{\providericonfile{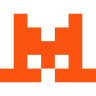}}

\newcommand{\providerMAPLE}{\inlineicon[1.1em]{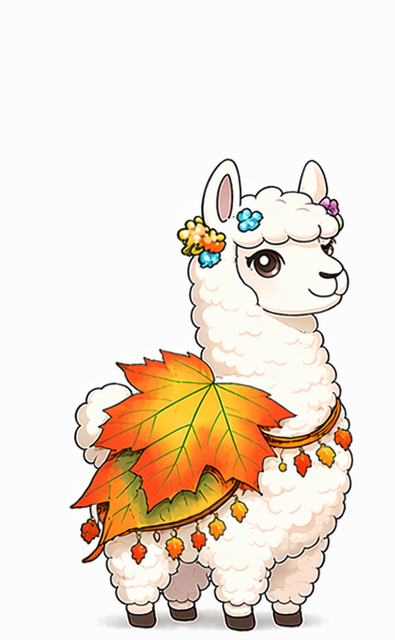}}

\newcommand{\topicGovernment}{\topiciconfile{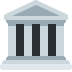}}
\newcommand{\topicCivic}{\topiciconfile{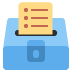}}
\newcommand{\topicEconomy}{\topiciconfile{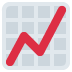}}
\newcommand{\topicEducation}{\topiciconfile{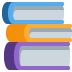}}
\newcommand{\topicMedia}{\topiciconfile{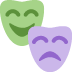}}
\newcommand{\topicSports}{\topiciconfile{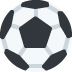}}
\newcommand{\topicPublicLife}{\topiciconfile{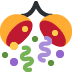}}
\newcommand{\topicGeography}{\topiciconfile{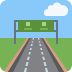}}

\newcommand{\flagUS}{\flagiconfile{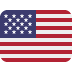}}
\newcommand{\flagCA}{\flagiconfile{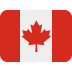}}
\newcommand{\flagJM}{\flagiconfile{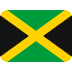}}
\newcommand{\flagIN}{\flagiconfile{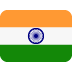}}
\newcommand{\flagPK}{\flagiconfile{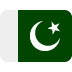}}
\newcommand{\flagBD}{\flagiconfile{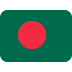}}
\newcommand{\flagLK}{\flagiconfile{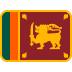}}
\newcommand{\flagHK}{\flagiconfile{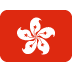}}
\newcommand{\flagMY}{\flagiconfile{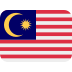}}
\newcommand{\flagPH}{\flagiconfile{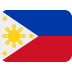}}
\newcommand{\flagSG}{\flagiconfile{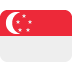}}
\newcommand{\flagNG}{\flagiconfile{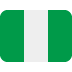}}
\newcommand{\flagZA}{\flagiconfile{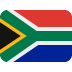}}
\newcommand{\flagKE}{\flagiconfile{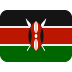}}
\newcommand{\flagGH}{\flagiconfile{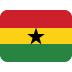}}
\newcommand{\flagTZ}{\flagiconfile{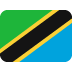}}
\newcommand{\flagGB}{\flagiconfile{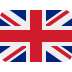}}
\newcommand{\flagIE}{\flagiconfile{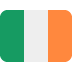}}

\newcommand{\LocalNewsQATopicLegend}{%
  \topicGovernment{}~politics/government, %
  \topicCivic{}~civic institutions, %
  \topicEconomy{}~economy/business, %
  \topicEducation{}~education, %
  \topicMedia{}~media/culture, %
  \topicSports{}~sports, %
  \topicPublicLife{}~public life/holidays, and %
  \topicGeography{}~geography/infrastructure%
}

\newcommand{\cmark}{%
\tikz[baseline=-0.6ex]\node[draw=none, fill=green!20, circle, inner sep=1pt] {\textcolor{green!60!black}{\checkmark}};}

\newcommand{\xmark}{%
\tikz[baseline=-0.6ex]\node[draw=none, fill=red!20, circle, inner sep=1pt] {\textcolor{red!70!black}{$\times$}};}

\def \llamachat{\textsc{LLaMA-3.2-1B-Instruct}}

\definecolor{localwithtestwith}{RGB}{163, 206, 168}
\definecolor{localwithtestwithout}{RGB}{236, 167, 164}
\definecolor{localwithouttestwith}{RGB}{250, 217, 183}
\definecolor{localwithouttestwithout}{RGB}{217, 217, 217}

\definecolor{globalwithtestwith}{RGB}{171, 205, 226}
\definecolor{globalwithouttestwithout}{RGB}{127, 127, 127}

\def \now{\textsc{Now}}
\newcommand{\Tplus}{\texttt{T+}}
\newcommand{\Tminus}{\texttt{T-}}
\newcommand{\Iplus}{\texttt{I+}}
\newcommand{\Iminus}{\texttt{I-}}

\def \localwith{\highlightRoundedBorder{localwithtestwith}{\textsc{Local} (\Tplus, \Iplus)}}
\def \localA{\highlightRounded{localwithtestwithout}{\textsc{Local} (\Tplus, \Iminus)}}
\def \localB{\highlightRoundedBorder{localwithouttestwith}{\textsc{Local} (\Tminus, \Iplus)}}
\def \localwithout{\highlightRounded{localwithouttestwithout}{\textsc{Local} (\Tminus, \Iminus)}}

\def \globalwith{\highlightRoundedBorder{globalwithtestwith}{\textsc{Global} (\Tplus, \Iplus)}}

\def \globalwithout{\highlightRounded{globalwithouttestwithout}{\textsc{Global} (\Tminus, \Iminus)}}

\begin{document}
\maketitle
\begin{abstract}
Large language models can memorize competing locale-specific facts yet fail to select among them when the locale changes, defaulting instead to a single globally dominant answer. We formalize this as \emph{localized knowledge disambiguation} and introduce \textsc{LocalNewsQA}, an $18{,}700$-item English-news benchmark that pairs the same question across two locales and scores whether a model actually switches its answer when the locale changes. We also introduce \textsc{MAPLE}, a controlled family of decoder-only models pretrained with document-level geographic metadata (source URL, country, and continent) already present in the training corpus, and compare it to metadata-free controls trained on identical data with the same token budget, architecture, and optimization. In controlled experiments at $1$B and $3$B, with inference-time metadata fixed, pretraining with metadata in \textsc{MAPLE} produces measurable switching and improves accuracy on questions whose correct answer depends on locale. Ablations and external-benchmark evaluations further suggest that locale-conditioned prediction benefits from geographic provenance learned during pretraining and that these benefits strengthen at larger model sizes.\footnote{Code is available at \url{https://github.com/iamshnoo/metadata_localization}}
\end{abstract}

\begin{figure}[t]
    \centering
    \includegraphics[width=\linewidth]{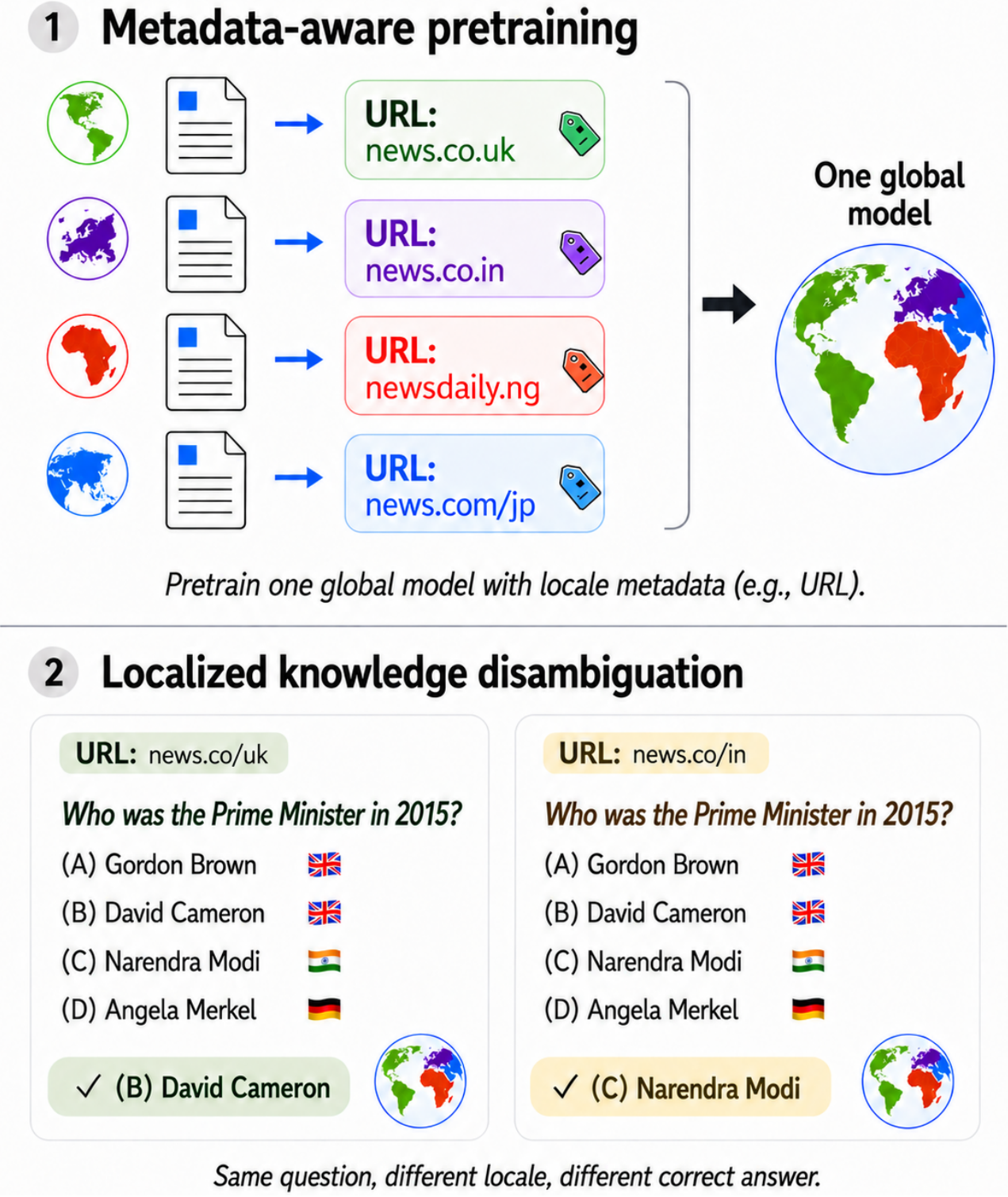}
    \caption{\textsc{MAPLE} turns geographic metadata already present in the training corpus into an inference-time locale signal, allowing one global model to condition factual predictions on source URL, country, and continent.}
    \label{fig:figure1}
\end{figure}

\section{Introduction}
\label{sec:introduction}
Ask a language model ``When does the school year start?'' and it will answer September, which is correct in the United States and wrong in Australia, India, Japan, and most of the Southern Hemisphere. The model may not be missing knowledge. It has likely memorized competing facts from many countries but has never been taught which fact belongs where.

This pattern is general. Large language models are trained on text from many locales yet optimized as a single broad distribution, collapsing locally valid alternatives into geographically homogenized defaults \citep{weidinger-2022-western,ghosh-2024-western}. ``What were the highlights of this weekend's football?'' has a different correct answer depending on whether the user is in New York or London, where the word refers to two different sports. ``Which agency releases census numbers?'' resolves to different institutions by country.
``Wrong'' answers in such cases are not recall failures. The model likely knows the correct answers but cannot select among them when the locale changes.

We call this problem \emph{localized knowledge disambiguation}, the task of selecting among locale-specific answer distributions rather than producing one unconditioned answer. It draws on under-specified question answering \citep{parrish-2022-bbq}, geographic QA \citep{yin-2022-geomlama,hofmann-2024-geographic}, cultural evaluation \citep{adilazuarda-2024-towards,choenni-2024-echoes}, and geography-aware adaptation \citep{cao-2025-specializing}, but isolates a stricter requirement, namely that the same question frame must resolve differently when the locale changes.

Existing geographic and cultural evaluations score isolated facts, opinions, values, norms, or situated tasks. They are necessary, but they do not directly test whether a model changes its answer under a locale change. Target-locale accuracy alone is insufficient because a model can answer one locale correctly while producing the same answer for a contrast locale. To our knowledge, no existing benchmark scores paired locale switching directly. On the method side, most localization work intervenes through continued pretraining, prompting, adaptation, or evaluation-time context \citep{gururangan-2020-dont,adilazuarda-2024-towards,cao-2025-specializing,hofmann-2024-geographic}. Whether document provenance available in training corpora can become a learned signal for locale-conditioned prediction, under controlled comparisons, remains untested.

We address both gaps. We introduce \textsc{LocalNewsQA}, an English-news benchmark that pairs the same question across two locales with different gold answers, making paired locale switching directly observable. We also introduce \textsc{MAPLE}, a family of decoder-only models trained from scratch with document-level geographic metadata (source URL, country, and continent) from the \now{} corpus \citep{davies-2016-now}. Because the benchmark questions are generated and validated independently of the training corpus, and the pretraining comparison holds everything except metadata fixed, improvements on \textsc{LocalNewsQA} can be attributed to the metadata intervention. Our study asks whether that single intervention changes a specific, directly measurable behavior. Our paper makes four contributions:

\squishlist
\item \textbf{Task and benchmark.} We define localized knowledge disambiguation and introduce \textsc{LocalNewsQA}, an $18{,}700$-item English-news benchmark that pairs the same question across two locales with different gold answers, enabling direct measurement of paired locale switching.
\item \textbf{Controlled pretraining.} We introduce \textsc{MAPLE}, models equipped to handle document-level geographic metadata (URL, country, continent), tested while holding data, architecture, and token budget fixed.
\item \textbf{Efficient localization.} With inference-time metadata held fixed, metadata-conditioned pretraining improves locale switching, with gains increasing from $1$B to $3$B.
\item \textbf{Mechanism and transfer.} Ablations, comparisons that cross training-time and inference-time metadata, and external benchmarks suggest that the full effect requires geographic provenance available during both pretraining and inference, and that it transfers to held-out evaluations.
\squishend

\section{Related Work}
\label{sec:related-work}
Localized knowledge disambiguation intersects under-specified question answering \citep{parrish-2022-bbq}, locale-sensitive prompting \citep{cao-2025-specializing}, culturally grounded evaluation \citep{adilazuarda-2024-towards,choenni-2024-echoes}, and geographic QA \citep{yin-2022-geomlama,hofmann-2024-geographic}. These works show that models encode uneven geographic and cultural assumptions, but they do not directly test whether a model changes its answer when the locale changes. \textsc{LocalNewsQA} targets that stricter question by pairing the same English question frame across two locales and scoring whether the model switches.

\textsc{XLQA}~\citep{roh-etal-2025-xlqa} and multilingual ODQA benchmarks~\citep{lewis-etal-2020-mlqa,artetxe-etal-2020-cross,clark-etal-2020-tydi,longpre-etal-2021-mkqa} share the premise that answers need not be locale-invariant, but they vary the language alongside the locale, making it difficult to attribute answer changes to geographic knowledge rather than translation quality or multilingual transfer. We hold language fixed (English throughout), train metadata-conditioned and metadata-free models from scratch, and score whether the model switches its answer when only the locale changes. This design isolates locale-conditioned answer selection from translation quality, multilingual transfer, and evaluation-time prompting.

Metadata conditioning has steered models with control codes \citep{keskar-2019-ctrl}, temporal signals \citep{dhingra-2022-time}, language tags \citep{liu-2020-multilingualdenoising}, document structure \citep{aghajanyan-2022-htlm}, human-preference labels \citep{korbak-2023-pretraininglanguagemodelshuman}, source attribution \citep{khalifa-2024-sourceawaretrainingenablesknowledge,weller-2024-according}, and social context \citep{kulkarni-etal-2021-lmsoc-approach,li-etal-2022-ntulm}. Most relevant to our work, \citet{gao-2025-metadataconditioningaccelerateslanguage} show that real URL metadata accelerates decoder-only pretraining, but they evaluate the effect through perplexity rather than testing whether the model can use provenance to disambiguate competing locale-specific answers at inference time. \textsc{MAPLE} addresses that open question, measuring whether geographic metadata becomes a conditioning signal for locale-sensitive answer selection through paired locale switching rather than perplexity alone.

\section{Task and Benchmark}
\label{sec:main}
Existing evaluations cannot measure whether a model changes its answer when the locale changes, because they test one locale at a time. We design a task and benchmark that makes paired locale switching directly observable.

\subsection{Task Definition}
\label{sec:localnewsqa-main}
Localized knowledge disambiguation evaluates whether a model can condition a factual answer on the locale where the fact applies. The key difficulty is that the question need not name the locale. Prompts such as ``Which agency releases census numbers?'', ``Which grades are primary school?'', or ``Which football league is the top flight?'' are ambiguous unless the locale is supplied, because the relevant statistical office, grade span, or sports code changes by country. The diagnostic examples are therefore not merely country-specific questions, but questions whose answer changes when the locale changes.

\textsc{LocalNewsQA} is designed around that requirement. The benchmark contains two evaluation splits. The \textit{explicit} split names the locale in the question and measures localized factual recall; these questions have a single correct answer once the locale is specified. The \textit{ambiguous} split omits the locale from the question text and supplies the locale through the evaluation context; these are locale-sensitive because the correct answer changes with the locale. Each ambiguous item also has a paired contrast locale with a different gold answer. This pairing makes \textsc{LocalNewsQA} a direct evaluation of the behavior that accuracy-only benchmarks miss, namely whether the same model changes its answer when the locale changes.

\subsection{\textsc{LocalNewsQA} Construction and Validation}
\label{sec:localnewsqa-construction-main}
We construct \textsc{LocalNewsQA} in English news because news provides dense local institutions, public bodies, sports, education systems, media references, and recent events with traceable source URLs. The benchmark has to satisfy broad country/topic coverage while keeping only rows whose answer can be verified against source evidence. We therefore use generation only to find candidate question frames, and filter to keep only rows that pass split-specific source and formatting checks. The final release contains $18{,}700$ questions from $17$ countries across $4$ continents, with $17{,}000$ explicit items balanced at $1{,}000$ per country and $1{,}700$ ambiguous items balanced at $100$ per country. The benchmark covers $8$ topic families, \LocalNewsQATopicLegend. Table~\ref{tab:localnewsqa_gold_pipeline} summarizes the gold-selection criteria.

\begin{table}[t]
\centering
\footnotesize
\setlength{\tabcolsep}{2.5pt}
\renewcommand{\arraystretch}{1.08}
\begin{tabular}{@{}lL{0.38\linewidth}L{0.38\linewidth}@{}}
\toprule
\textbf{Criterion} & \textbf{Explicit split} & \textbf{Ambiguous split} \\
\midrule
Balance & $1{,}000$ items per country; $8$ topic families represented & $100$ items per country; $8$ topic families represented \\
Format & $4$ options, $1$ gold, no duplicate question & $4$ options, $1$ gold, no duplicate question \\
Evidence & Target answer supported by source & Both target and contrast answers supported \\
Locale dep. & Question names the locale & Question omits locale; target and contrast answers differ \\
\bottomrule
\end{tabular}
\caption{Gold-selection criteria for the final \textsc{LocalNewsQA} benchmark. Questions that fail any criterion are excluded.}
\label{tab:localnewsqa_gold_pipeline}
\end{table}
Every question in the final benchmark includes source evidence for auditability. In a three-annotator audit of a stratified $100$-item gold sample, majority vote accepted $96/100$ items, including $50/50$ explicit and $46/50$ ambiguous, with $98.0\%$ average pairwise agreement and Fleiss' $\kappa=0.76$ on the binary accept/reject decision. Appendix~\ref{sec:localnewsqa-appendix} gives the full construction, validation, and human-audit protocol.

\subsection{\textsc{LocalNewsQA} Evaluation}
\label{sec:localnewsqa-eval-main}
We report target-locale accuracy and paired locale switching. For an ambiguous pair with frame $q$, target locale $\tgt{\ell_t}$, contrast locale $\ctr{\ell_c}$, and answers $\tgt{a_t},\ctr{a_c}$, target accuracy asks only whether the model chooses $\tgt{a_t}$ under $\tgt{\ell_t}$. The paired metrics are defined as follows.
\[
\begin{aligned}
\textit{exact pair}\quad &\underbrace{\hat a(q,\tgt{\ell_t})=\tgt{a_t}}_{\text{correct on target}} \;\land\; \underbrace{\hat a(q,\ctr{\ell_c})=\ctr{a_c}}_{\text{correct on contrast}},\\[6pt]
\textit{margin switch}\quad
&\underbrace{s(\tgt{a_t}\mid q,\tgt{\ell_t})>s(\ctr{a_c}\mid q,\tgt{\ell_t})}_{\text{target ans.\ preferred under }\tgt{\ell_t}}\\[6pt]
&{}\land\; \underbrace{s(\ctr{a_c}\mid q,\ctr{\ell_c})>s(\tgt{a_t}\mid q,\ctr{\ell_c})}_{\text{contrast ans.\ preferred under }\ctr{\ell_c}}.
\end{aligned}
\]
Exact pair is stringent because both sides must be correct; margin switch is softer but still requires the score ordering to move toward the locale-appropriate answer on both sides.

For \textsc{MAPLE}, we score pretrained checkpoints using length-normalized log-probability over each option's token sequence. Table~\ref{tab:qa-factorial} averages the crossed paired-switch comparison over five answer-order seeds; target-accuracy and other analysis fix a shared answer-order seed and use question-level bootstrap intervals. Metadata-formatted inference supplies the locale fields corresponding to the question being evaluated. Target prompts include the target country and continent, and contrast prompts include the contrast country and continent. Figure~\ref{fig:figure9_prompts} gives the exact prompts, and Figure~\ref{fig:qa-appendix} reports results on the chat variants obtained from LoRA instruction tuning.

\section{\textsc{MAPLE} Controlled Pretraining}
\label{sec:maple-main}
We use the \now{} corpus \cite{davies-2016-now} throughout. Each document includes a publication year, source URL, and country of origin, which we map to continents and serialize as structured metadata before the article title and content. After filtering, the corpus contains approximately $4$M documents from Africa, $10$M from America, $7.5$M from Asia, and $5.6$M from Europe. All local and global models are trained on $41.9$B tokens. Local models sample one continent to that budget, while global models mix all four continents before sampling. Appendix~\ref{sec:hyperparam} gives the full data formatting and geographic coverage details.

\begin{table}[t]
\centering
\scriptsize
\begin{tabular}{@{}cccc@{}}
\toprule
\textbf{\# Continents} & \textbf{Train Meta} & \textbf{Eval Meta} & \textbf{Notation} \\
\midrule
1  & \cmark\,\Tplus & \cmark\,\Iplus & \localwith \tabstrut\\
1  & \cmark\,\Tplus & \xmark\,\Iminus & \localA \tabstrut\\
1  & \xmark\,\Tminus & \cmark\,\Iplus & \localB \tabstrut\\
1  & \xmark\,\Tminus & \xmark\,\Iminus & \localwithout \tabstrut\\
\midrule
4 & \cmark\,\Tplus & \cmark\,\Iplus & \globalwith \tabstrut\\
4 & \xmark\,\Tminus & \xmark\,\Iminus & \globalwithout \tabstrut\\
\bottomrule
\end{tabular}
\caption{Model variants used in this paper. ``\texttt{T+/T-}'' indicate metadata present/absent during training; similarly ``\texttt{I+/I-}'' indicate metadata available versus not available at inference. \highlightRoundedBorder{white}{Bordered boxes} also indicate \texttt{I+} models.}
\label{tab:model_variants}
\end{table}

\begin{center}
\begin{tcolorbox}[
    enhanced,
    colback=blue!3!white,
    colframe=black!70,
    coltitle=black,
    colbacktitle=blue!10!white,
    title=\centering\textbf{Metadata-formatted training instance},
    fontupper=\ttfamily\footnotesize,
    fonttitle=\normalsize,
    width=\linewidth,
    boxrule=0.35mm,
    arc=1mm,
    top=0.7mm, bottom=0.7mm,
    left=1mm, right=1mm,
    boxsep=1.5mm
]
\textbf{URL}\quad \texttt{news.example/jp/article/12345}\\
\textbf{COUNTRY}\quad Japan\\
\textbf{CONTINENT}\quad Asia\\[0.25em]
\textbf{TITLE}\quad Election results spark debate \ldots\\[0.25em]
\textbf{CONTENT}\quad The recent election results have led to widespread discussions across political parties and the public, with analysts noting \ldots
\end{tcolorbox}
\end{center}

The pretraining intervention changes only the document serialization. Metadata-formatted documents prepend source URL, country, and continent fields before the title and article text, as shown above. Metadata-free documents omit these fields. The comparison is controlled so that changes in behavior can be attributed to the presence of provenance metadata rather than to data volume, architecture, optimizer, or tokenizer changes.

We train $33$ models from scratch at two sizes. At $500$M, we train $10$ models covering $4$ continents $\times$ $2$ metadata settings plus $2$ global settings. At $1$B, we train $23$ models with the same $10$ settings plus $5$ provenance-granularity ablations and $8$ leave-one-continent controls. We additionally train two $3$B global models to test scaling.

Throughout the results, \Tplus/\Tminus{} indicate metadata present or absent during training, and \Iplus/\Iminus{} indicate metadata present or absent at inference. The notation separates whether a model has learned from metadata during pretraining from whether metadata is supplied at evaluation time.

\section{Results}
Each research question builds on the previous one, moving from establishing the basic effect to explaining which signals drive it and whether it generalizes. RQ1 (\S\ref{sec:exp4}) establishes paired locale switching. RQ2 (\S\ref{sec:exp1}) tests whether metadata also improves region-specific language modeling. RQ3 (\S\ref{sec:exp3}) isolates which provenance fields carry the signal. RQ4 (\S\ref{sec:external-main}) checks whether gains transfer to external benchmarks.

\subsection{RQ1. Does \textsc{MAPLE} Improve Localized Knowledge Disambiguation?}
\label{sec:exp4}
The first experiment asks whether provenance metadata improves localized disambiguation under a controlled pretraining intervention. If the model learns metadata as a usable conditioning signal, it should help most when the question underspecifies locale, producing larger accuracy gains on ambiguous items than on explicit ones. Paired target/contrast cases provide the strongest test because the answer must change with the locale, and switching gains should increase at larger model sizes if the metadata signal is being used more effectively.

\paragraph{Results}
\looseness=-1 The primary comparison controls inference-time information. Both $(\Tplus,\Iplus)$ and $(\Tminus,\Iplus)$ receive different locale prompts, but only the former learned from metadata during pretraining. Table~\ref{tab:qa-factorial} shows that pretraining raises the paired margin-switch rate from $6.04\%$ to $7.78\%$ at $1$B ($+1.74$ points) and from $21.36\%$ to $24.13\%$ at $3$B ($+2.77$ points). These increments isolate the effect attributable to metadata-conditioned pretraining.

\begin{insightbox}
\textbf{Takeaway 1.} With inference-time metadata fixed, metadata-conditioned pretraining improves paired-switching, with increasing gains from $1$B to $3$B.
\end{insightbox}

\begin{figure}[t]
    \centering
    \includegraphics[width=\linewidth]{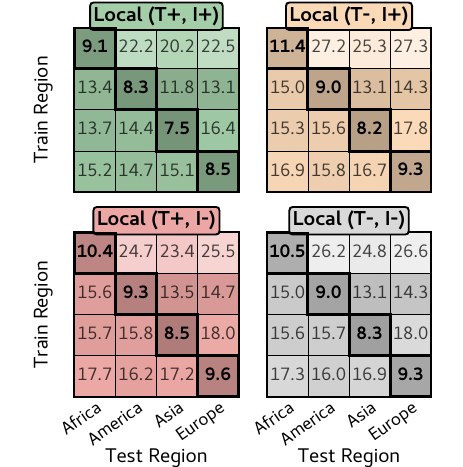}
    \caption{\textbf{Cross-continent perplexity for local models.} Diagonal cells are consistently lower than off-diagonal cells, showing locality. Full metadata gives the strongest local performance and improves each train/evaluation setting compared with weaker metadata variants.}
    \label{fig:local-crosstest}
\end{figure}

\begin{table}[t]
\centering
\small
\setlength{\tabcolsep}{3.2pt}
\begin{tabular}{@{}lccccc@{}}
\toprule
Scale & $(\Tminus,\Iminus)$ & $(\Tplus,\Iminus)$ & $(\Tminus,\Iplus)$ & $(\Tplus,\Iplus)$ & $\Delta_{\mathrm{train}}$ \\
\midrule
$1$B & $0.00$ & $0.00$ & $6.04$ & $\mathbf{7.78}$ & $\mathbf{+1.74}$ \\
$3$B & $0.00$ & $0.00$ & $21.36$ & $\mathbf{24.13}$ & $\mathbf{+2.77}$ \\
\bottomrule
\end{tabular}
\caption{Paired margin-switch rate (\%) on the $1{,}700$ ambiguous \textsc{LocalNewsQA} pairs, averaged over five answer-order seeds. The primary pretraining contrast, $(\Tplus,\Iplus)$ vs. $(\Tminus,\Iplus)$, holds inference-time locale information fixed; $\Delta_{\mathrm{train}}$ is their difference in percentage points. Both $\Iminus$ settings use identical target/contrast prompts and therefore yield zero switches by construction; $(\Tminus,\Iminus)$ is the negative control.}
\label{tab:qa-factorial}
\end{table}

\looseness=-1 The $(\Tminus,\Iminus)$ zero paired-switch result is expected because target and contrast prompts are identical; we include it only as a negative control showing that the evaluation does not leak locale information.

In our additional target-locale accuracy analysis, we compare
$(\Tplus,\Iplus)$ against $(\Tminus,\Iminus)$. On the ambiguous
split, this comparison yields gains of $+6.7$ points at $1$B
and $+9.2$ points at $3$B. The explicit split also improves, but less sharply, because explicit questions already name the locale. These gains are distributed across countries and topics, and their bootstrap intervals exclude zero at both model sizes (Appendix Table~\ref{tab:localnewsqa_model_gain_summary}).

\subsection{RQ2. Does Metadata Improve Region-Specific Language Modeling?}
\label{sec:exp1}
\label{sec:exp2}

\looseness=-1 If metadata helps the model condition on locale, the benefit should extend beyond the benchmark format and also lower held-out perplexity on text from the same geographic region. We test this at two levels. First, continent-specific local models, where crossed train/evaluation settings separate learned provenance from metadata supplied only at inference. Second, global models trained on all continents, where the question is whether one model can retain multiple local distributions without region-specific parameters.

The four crossed settings are abbreviated as \localwith{} (metadata during both training and inference), \localA{} (train-only), \localB{} (inference-only), and \localwithout{} (no metadata at either stage).

\paragraph{Setup} For each continent and scale ($0.5$B and $1$B), we train checkpoints with and without metadata and evaluate under all four crossed settings on local and cross-continent test sets (Figure~\ref{fig:local-crosstest}). Each continent has a held-out test set of $1{,}000$ documents, and all models are trained on the same total number of tokens. Test perplexity is computed over non-metadata tokens only, so any improvement reflects better content prediction rather than metadata memorization. We additionally train global models on data from all continents at $0.5$B, $1$B, and $3$B, asking whether one model can serve multiple regions without region-specific specialization. The global test set contains $1{,}000$ documents with $250$ sampled uniformly from each continent, testing whether a single set of parameters can serve all regions without sacrificing performance on any individual one.

\paragraph{Hypotheses} If metadata provides a useful conditioning signal, models trained with it should achieve lower perplexity on their continent's test set than metadata-free controls, with the crossed settings (metadata at only one stage) falling between the two extremes. Local models should remain local, with lower same-continent than cross-continent perplexities. At the global level, a metadata-conditioned model should approach local-model quality on each region while staying stronger on average on the combined global test set.
\begin{figure}[t]
    \centering
    \includegraphics[width=\linewidth]{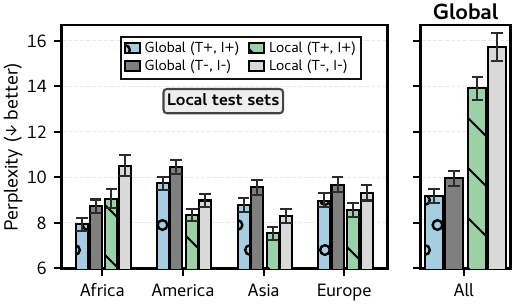}
    \caption{\textbf{Global metadata conditioning improves held-out perplexity at $1$B.} A single global model with metadata lowers perplexity on every continent-specific test set and the combined global test set.}
    \label{fig:local-global}
\end{figure}

\paragraph{Results}
\looseness=-1 Figure~\ref{fig:local-crosstest} shows that models trained and evaluated with full metadata achieve the lowest perplexity on every same-continent test set, showing that metadata indeed provide a useful conditioning signal. The crossed settings fall in between; a supporting diagnostic in Appendix Figure~\ref{fig:local-crosstest-full} suggests that inference-time metadata contributes more than train-time metadata alone. Across all configurations, same-continent perplexity is consistently lower than cross-continent perplexity, showing that local models remain local.

Figure~\ref{fig:local-global} summarizes the global comparison at $1$B. A single metadata-conditioned global model lowers perplexity relative to the metadata-free control on every continent-specific test set as well as on the combined global test set. From $500$M to $1$B, it approaches local-model quality while remaining stronger on the combined global test set. The remaining gap is expected because region-specific models specialize to one distribution, whereas the global model preserves all four under one parameter budget. The useful result is that metadata moves one shared model toward local behavior without sacrificing breadth. This pattern continues at $3$B, where the metadata-conditioned global model again outperforms the metadata-free control. The training-token curves in Figure~\ref{fig:token-efficiency} show that metadata-conditioned models reach any given perplexity target using fewer training tokens, an efficiency benefit that persists across model sizes.

\begin{insightbox}
\textbf{Takeaway 2.} A single global model can retain multiple regional distributions when trained with provenance metadata. The benefit persists from $500$M to $3$B and appears as both lower perplexity and faster convergence.
\end{insightbox}

We additionally observe that localized performance is distributed across countries rather than carried by a single dominant country per continent, and that Africa-trained local models transfer less symmetrically to other regions (Appendix Figures~\ref{fig:local-country} and \ref{fig:local-asymmetry}).

\begin{figure*}[t]
    \centering
    \includegraphics[width=0.9\textwidth]{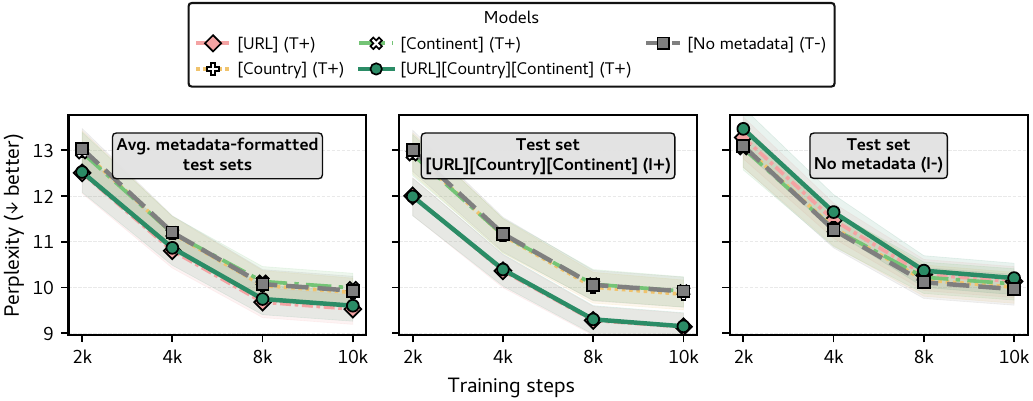}
    \caption{\textbf{Metadata ablations across checkpoints.} \URLChip{}-only metadata recovers most of the full-metadata perplexity benefit; \CountryChip{}/\ContinentChip{} tags alone trail \URLChip{}, and adding them to \URLChip{} changes little.}
    \label{fig:ablation-meta}
\end{figure*}

\subsection{RQ3. Which Provenance Signals Matter, and What Can They Not Replace?}
\label{sec:exp3}
This experiment isolates which provenance fields carry the signal and whether metadata can compensate for missing regional coverage.

\paragraph{Setup} We evaluate five $1$B metadata-granularity variants, \URLChip{}, \URLCountryChip{}, \URLContinentChip{}, \CountryChip{}, and \ContinentChip{}, alongside full/no metadata and leave-one-continent global controls. Each variant uses the same training data and token budget, differing only in which metadata fields are serialized before the document text. Figure~\ref{fig:ablation-meta} reports the main variants; Appendix Figures~\ref{fig:ablation-meta-all} and \ref{fig:ablation-leaveone} show all checkpoints and the leave-one-continent results.

\looseness=-1 Leave-one-continent controls separate metadata from data coverage. Each model is trained on three of the four continents and evaluated on both the held-out continent and the combined global test set. This design tests a specific alternative explanation. If metadata merely encoded continent labels, adding it should compensate for the missing region even when that region was excluded from training. If, on the other hand, metadata helps the model organize evidence it has actually seen, held-out-region perplexity should still degrade when the region was absent from training, regardless of what metadata is supplied at evaluation time. The distinction matters for practitioners deciding whether metadata can substitute for broader data collection.

\paragraph{Hypotheses} \looseness=-1 If provenance is the key signal, the URL field alone should recover most of the benefit because URLs implicitly encode geography through country-code top-level domains and region-specific domain names (Appendix Table~\ref{tab:url-signals} quantifies this). Coarser labels such as country or continent names should be less informative in isolation, and metadata of any granularity should not compensate for data the model has never seen during training.

\begin{figure*}[t]
    \centering
    \captionsetup[subfigure]{font=footnotesize,skip=4pt}
    \begin{subfigure}[t]{0.49\textwidth}
        \centering
        \includegraphics[width=\linewidth]{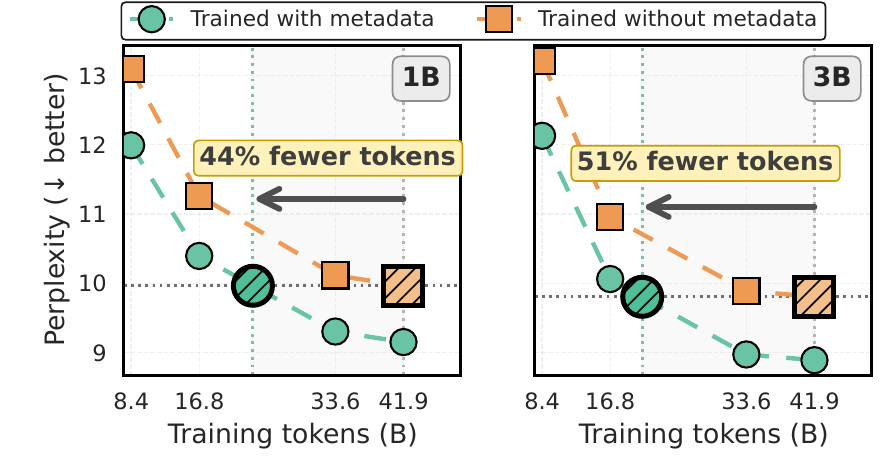}
        \caption{Metadata reaches target perplexity sooner.}
        \label{fig:token-efficiency}
    \end{subfigure}
    \hfill
    \begin{subfigure}[t]{0.49\textwidth}
        \centering
        \includegraphics[width=0.92\linewidth]{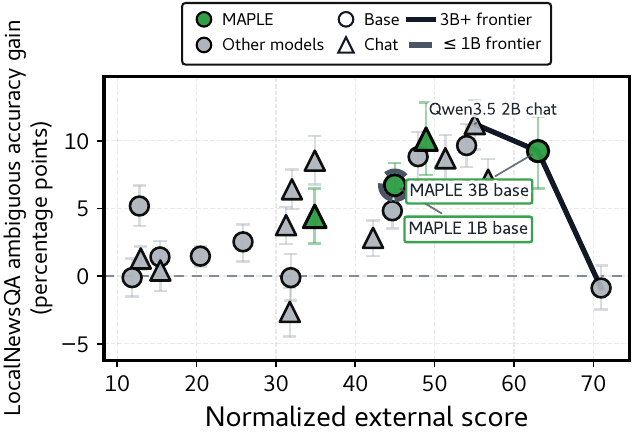}
        \caption{Localization gains preserve external transfer.}
        \label{fig:downstream-gain-pareto}
    \end{subfigure}
    \vspace{0.45em}
    \caption{\textbf{Scale and transfer.} Metadata lowers global perplexity more quickly from $1$B to $3$B (lower is better in panel A) while preserving broad external performance. In panel B, each point is a model; the frontier is the set of models not dominated by any other model of similar size on both axes. \textsc{MAPLE} points remain on this frontier. Task-level external results are in Appendix Table~\ref{tab:external_eval_summary}.}
    \label{fig:scale-transfer}
\end{figure*}

\paragraph{Results} The results support all three expectations. Figure~\ref{fig:ablation-meta} shows that \URLChip{} is the strongest individual provenance signal, nearly matching full metadata and substantially outperforming the no-metadata baseline. The URL field alone recovers most of the localization benefit. \CountryChip{} and \ContinentChip{} trail \URLChip{} and add little when paired with it, suggesting that coarse geographic labels are largely redundant once the URL is present. A supporting leave-one-continent analysis (Appendix Figure~\ref{fig:ablation-leaveone}) shows that excluding any continent raises held-out and global perplexity regardless of metadata, so provenance organizes evidence the model has seen but cannot substitute for regional coverage it lacks. This finding has a practical implication: metadata conditioning and broad geographic data collection serve different purposes and neither replaces the other.

\begin{insightbox}
\textbf{Takeaway 3.} The source URL alone carries most of the geographic signal. Adding country or continent labels on top of it changes little, and no amount of metadata can teach the model about regions it has never seen in training.
\end{insightbox}

\subsection{RQ4. Does Localization Transfer Beyond \textsc{LocalNewsQA}?}
\label{sec:external-main}
Finally, we test whether localized disambiguation gains transfer beyond \textsc{LocalNewsQA}. We evaluate six benchmarks covering geographic factual recall (\textsc{GeoMLaMA}; \citealp{yin-2022-geomlama}), opinions and values (\textsc{GlobalOpinionQA}; \citealp{durmus-2024-globalopinionqa}, \textsc{WorldValuesBench}; \citealp{choenni-2024-echoes}), norm judgments (\textsc{NormAD}; \citealp{rao-2025-normad}), and culturally situated tasks (\textsc{BLEnD}; \citealp{myung-2024-blend}, \textsc{Global-MMLU-CS}; \citealp{singh-2025-globalmmlu}). Figure~\ref{fig:downstream-gain-pareto} plots each model's \textsc{LocalNewsQA} ambiguous gain against its normalized external score (the mean of per-task accuracy scores normalized by the best model on each task). The question is whether training with metadata to improve localized disambiguation hurts or helps performance on broader geographic and cultural evaluations.

\paragraph{Hypotheses} A useful localization method should improve controlled disambiguation without sacrificing broad performance on external evaluations. Inference-time metadata alone, without a matching pretraining signal, is not expected to produce the same benefit as reliably.

\paragraph{Results} Figure~\ref{fig:downstream-gain-pareto} shows that the controlled \textsc{MAPLE} points remain competitive with size-matched models on normalized external score while simultaneously improving \textsc{LocalNewsQA}. At $3$B, \textsc{MAPLE} shows statistically significant gains on \textsc{GlobalOpinionQA} ($+8.4$), \textsc{NormAD} ($+5.1$), and \textsc{BLEnD} ($+13.5$), with positive point estimates on the remaining three tasks (full results in Appendix Table~\ref{tab:external_eval_summary}). Baseline models with metadata appended only at evaluation time show mixed gains, consistent with the interpretation that locale metadata is more reliable when learned during pretraining than when supplied only at inference.

\begin{insightbox}
\textbf{Takeaway 4.} On six external benchmarks, \textsc{MAPLE} remains competitive with size-matched models while improving locale-conditioned behavior.
\end{insightbox}

\section{Discussion}
\label{sec:discussion}
\looseness=-1 Taken together, the paired-switching, perplexity, and transfer results suggest that, at least for news-domain text at the scales tested, preserving provenance during pretraining adds a modest but consistent benefit beyond inference-time metadata alone. In practice, this means geographic metadata already present in the training corpus should be preserved rather than stripped during preprocessing. The URL field likely carries most of the signal in part because it encodes country-code TLDs, region-specific domains, and outlet identity (Appendix Table~\ref{tab:url-signals}), all of which are lost when corpora are reduced to plain text.

\paragraph{Beyond geography} The paired-switching protocol is not inherently geographic. Any axis along which valid completions compete could be tested with the same factorial design. Temporal provenance (publication date) would let a model distinguish ``Who is the president?'' across time periods, and source-type provenance (books vs.\ web data) could help the model weight formal versus colloquial usage depending on context. The shared requirement is that competing valid answers exist and that an available metadata field can disambiguate among them.

\section{Conclusion}
\label{sec:conclusion}
\looseness=-1 Geographic metadata already present in training corpora can help a model bind facts to locales. We show that preserving source URLs, country, and continent fields from the NOW news corpus during pretraining gives a single global model the ability to condition its answers on locale. Holding inference-time locale information fixed, metadata-conditioned pretrained models are better at switching answers when the locale changes, with progressively increasing strength from $1$B to $3$B. The source URL alone carries most of the geographic signal without requiring explicit country or continent tags. These gains transfer to three of six external benchmarks with positive point estimates on the remaining three, suggesting that provenance-conditioned pretraining is a promising direction for locale-aware factual prediction.

\section*{Limitations}
\label{sec:limitations}
This study is intentionally designed as a controlled isolation of metadata conditioning rather than as an exhaustive survey of every setting where provenance could be useful. We fix the model family, tokenizer, optimizer, token budget, and source domain so that train-time provenance can be compared cleanly against metadata-free controls. The natural next step is broader transfer through the same factorial design across additional model families, multilingual corpora, open-web sources such as Common Crawl, and non-news domains where provenance may encode different kinds of context.

Several extensions would sharpen this picture further. Our from-scratch pretraining experiments cover $500$M to $3$B parameters; we did not test $70$B or larger models because that scale was infeasible under our academic compute budget. Whether the paired-switching gains persist beyond $3$B therefore remains untested, and we encourage future large-scale compute collaborations to evaluate this regime. Repeated pretraining runs would also separate evaluation uncertainty from pretraining stochasticity. \textsc{LocalNewsQA} is generated to target a specific behavior, paired locale switching, and is released with split-specific source validation plus a stratified human audit; future benchmark work can combine this controlled template with naturally occurring user questions and richer human annotation. These directions would test generality while preserving the main purpose of the present study, which is to isolate whether geographic metadata already present in training corpora can help a single global model bind facts to the locales where they are valid.

\section*{Ethical Considerations}
\label{sec:ethical}
Metadata conditioning could reinforce coarse geographic stereotypes if applied carelessly. Our study uses factual news content with verified provenance and avoids personal data. Geographic metadata should be treated as contextual signals rather than as proxies for individual identity or values. Downstream applications should consider fairness and inclusivity, particularly in high-stakes settings where locale-conditioned outputs could encode or amplify regional biases.

\section*{Acknowledgments}
\label{sec:ack}
We are thankful to the reviewers who provided feedback in earlier versions of this work. This work was generously supported by the US National Science Foundation CAREER award 2439202. This work is also in part supported by NSF grant IIS-2452129. We acknowledge the support by resources provided by the Office of Research Computing at George Mason University (https://orc.gmu.edu) and funded in part by grants from the NSF (Award Number 2018631). Any opinions, findings, and conclusions or recommendations expressed in this material are solely those of the authors.
\bibliography{anthology,custom}

\appendix
\clearpage
\section*{Appendix}
\label{sec:appendix}

We provide more details on the training process, hyperparameters, prompts, and additional results.

\renewcommand{\thesection}{A.\arabic{section}}
\section{Training details and Hyperparameters}
\label{sec:hyperparam}

\paragraph{Pretraining} We purchased an academic license for the \now{} corpus for $395$ USD. All model runs use a fixed random seed. We use the \textsc{LLaMA-3.2-1B} architecture \citep{grattafiori-2024-llama3} as the base design and set the vocabulary size to $128{,}256$. The only differences across the $0.5$B, $1$B, and $3$B parameter models are the number of layers, hidden size, number of heads, number of KV heads, and FFN size (Table~\ref{tab:model_configs}); all other configurations are shared.

We use a sequence length of $2048$. Our parameters for data parallelism, tensor parallelism, and pipeline parallelism are $4$, $1$ and $1$ respectively, so we effectively only use data parallelism. Our micro batch size is $8$, and we perform gradient accumulation every $64$ steps. Under these parameters, every single step corresponds to $\sim4$M tokens. On average, our GPU cluster is able to process $100{,}000$ tokens per second on a single node of $4$ GPUs, with $\sim185$ TFLOPs of processing power per GPU.

We train for $10{,}000$ total steps, validate on a held-out set of $1{,}000$ examples every $1{,}000$ steps, and save checkpoints at the same frequency. We use a linear warm-up from $0$ to $3e-3$ over $500$ steps, followed by cosine decay to $3e-4$ over the remaining $9{,}500$ steps. Although warm-up ratios of $5$-$10\%$ are more typical at larger scales, pilot runs on our data and hardware showed that shorter warm-up performed best, so we adopted it throughout. We selected $10{,}000$ total steps after extending pilot runs to $30{,}000$ steps and observing negligible further change in validation and test perplexity beyond the $10{,}000$-step mark.

We train with AdamW, weight decay $0.033$, and gradient clipping at $0.1$.

Each model is trained on a single node with four NVIDIA A100 GPUs (80GB each). The $500$M models take roughly $3$ days to train, and the $1$B models take roughly $5$ days. Across $10$ models at $500$M and $23$ models at $1$B, this amounts to approximately $145$ days of runtime.

\paragraph{Training data.} We use the \now{} corpus \cite{davies-2016-now} for all model studies. Each document includes a publication year, source URL, and country of origin, which we extract as metadata and map to continents using the United Nations geoscheme (Table~\ref{tab:continent_country_mapping}). Training documents are formatted by prepending structured metadata, namely URL, country, and continent, to the article title and content as shown in the main text example.

Training separate country-level models is not feasible in a controlled setting due to large disparities in data availability, so we train models at the continent level. After filtering, the corpus contains approximately $4$M documents from Africa, $10$M from America, $7.5$M from Asia, and $5.6$M from Europe. To control for data scale, all local and global models are trained on the same total of $41.9$B tokens, which is sufficient for the parameter scales we use \cite[DCLM;][]{li-2024-datacomplm}. For global models, data from all continents is combined in a randomized order before sampling to the token budget, while local models sample from a single continent to the same budget. This controlled setup ensures that performance differences arise from metadata conditioning and data composition rather than training scale.

\begin{table}[t]
\centering
\scriptsize
\begin{tabular}{@{}p{0.17\columnwidth}L{0.76\columnwidth}@{}}
\toprule
\textbf{Continent} & \textbf{Countries} \\
\midrule
America & \flagUS{}~United States, \flagCA{}~Canada, \flagJM{}~Jamaica \\
Asia & \flagIN{}~India, \flagPK{}~Pakistan, \flagBD{}~Bangladesh, \flagLK{}~Sri Lanka, \flagHK{}~Hong Kong, \flagMY{}~Malaysia, \flagPH{}~Philippines, \flagSG{}~Singapore \\
Africa & \flagNG{}~Nigeria, \flagZA{}~South Africa, \flagKE{}~Kenya, \flagGH{}~Ghana, \flagTZ{}~Tanzania \\
Europe & \flagGB{}~United Kingdom, \flagIE{}~Ireland \\
\bottomrule
\end{tabular}
\caption{\textbf{Geographic coverage of model-data countries.} Countries are grouped by continent. \textsc{LocalNewsQA} uses this coverage except for Singapore, yielding the $17$-country question set described in the main benchmark overview.}
\label{tab:continent_country_mapping}
\end{table}

\begin{table}[t]
\centering
\scriptsize
\begin{tabular}{@{}l@{}}
\toprule
\textbf{URL} \\
\midrule
\texttt{www.factquizmaster.com} \\
\texttt{www.globalfactcheck.org} \\
\texttt{www.worldknowledgehub.com} \\
\texttt{www.civicspedia.org} \\
\texttt{www.internationalfacts.net} \\
\texttt{www.currentaffairsdesk.com} \\
\texttt{www.newsinsightarchive.com} \\
\texttt{www.globalquizvault.com} \\
\texttt{www.factualdigest.org} \\
\texttt{www.publicknowledgebase.net} \\
\bottomrule
\end{tabular}
\caption{Synthetic base-domain pool used for metadata-formatted SFT and evaluation prompts.}
\label{tab:url_pool}
\end{table}

\begin{table}[t]
\centering
\scriptsize
\begin{tabular}{@{}cccccc@{}}
\toprule
\textbf{Layers} & \textbf{Hidden} & \textbf{Heads} & \textbf{KV Heads} & \textbf{FFN Size} & \textbf{\#Params} \\
\midrule
$24$ & $1024$ & $16$ & $16$ & $4096$ & $500$M \\
$16$ & $2048$ & $16$ & $16$ & $5632$ & $1$B \\
$28$ & $3072$ & $32$ & $32$ & $8192$ & $3$B \\
\bottomrule
\end{tabular}
\caption{Model architecture configurations.}
\label{tab:model_configs}
\end{table}

\paragraph{Architecture sanity check.}
To separate the metadata effect from the decoder family, we trained an additional matched $1$B pair with a Qwen2-style decoder through the final $10{,}000$-step checkpoint. This pair keeps the \textsc{LLaMA-3.2-1B} tokenizer, training corpus, optimizer, evaluated token budget, and metadata serialization fixed, and changes the transformer block implementation. Holding tokenization and vocabulary fixed is what makes this an architecture-only comparison rather than a joint change in architecture and text segmentation. Table~\ref{tab:qwen_arch_sanity} evaluates the two final checkpoints on $1{,}000$ held-out global documents under the same content-token perplexity protocol used for the main global models. The metadata-trained Qwen2-style checkpoint has lower perplexity than the metadata-free checkpoint on the metadata-formatted held-out set, and its matched-format score is also lower than the no-metadata matched-format score. This supports the interpretation that the provenance effect is not specific to the LLaMA-style decoder.

\begin{table}[t]
\centering
\scriptsize
\setlength{\tabcolsep}{2.5pt}
\begin{tabular}{@{}lcc@{}}
\toprule
\textbf{Checkpoint} & \textbf{Metadata test} & \textbf{No-metadata test} \\
\midrule
Trained with metadata & $9.33$ [$9.00$, $9.63$] & $10.45$ [$10.09$, $10.79$] \\
Trained without metadata & $9.85$ [$9.51$, $10.16$] & $9.97$ [$9.62$, $10.29$] \\
\bottomrule
\end{tabular}
\caption{\textbf{Architecture sanity check.} Held-out content-token perplexity for matched Qwen2-style $1$B checkpoints at the final $10{,}000$-step checkpoints. Lower is better. The pair uses the same \textsc{LLaMA-3.2-1B} tokenizer and data pipeline as the main controlled runs to isolate the decoder architecture.}
\label{tab:qwen_arch_sanity}
\end{table}

\section{SFT details}
\label{sec:sft-details}

This section reports \textsc{LocalNewsQA} results under a LoRA post-training protocol. We use LoRA adapters \citep{hu2022lora,dettmers-2023-qlora,schulman-2025-lora} to make post-training computationally efficient while leaving the pretrained base models frozen. We instruction-tune on the Alpaca dataset\footnote{\href{https://huggingface.co/datasets/yahma/alpaca-cleaned}{Hugging Face link for Alpaca data}} because its instruction format is close to our multiple-choice benchmark while exhibiting no content overlap with it. To preserve the pretraining format, we construct metadata-formatted and no-metadata SFT inputs that match the corresponding base models.

For metadata-formatted SFT instances, we sample base domains from the synthetic pool in Table~\ref{tab:url_pool}, add internally consistent country and continent tags, and then wrap the example with the \llamachat{} chat template (Figure~\ref{fig:sft_prompt}). Alpaca examples are not used as geographically grounded evidence; this SFT step only teaches the response format while preserving the metadata serialization seen during pretraining. At evaluation time, the base domain may be drawn from the same pool, but the country code, country, and continent are aligned to the corresponding evaluation example. This keeps target and contrast prompts grounded in their intended locales. For parameter-efficient finetuning, we use LoRA rank $256$, alpha $16$, and all linear layers as targets; this corresponds to approximately $161.5$M trainable adapter parameters for the $1$B models and $418.4$M for the $3$B models. We train for $3$ epochs with per-device batch size $2$, gradient accumulation $8$, learning rate $2e{-4}$, and the $8$-bit AdamW optimizer.

At evaluation time, we score the chat checkpoints on the full \textsc{LocalNewsQA} target-locale split using the same per-size prompt families as in the pretrained closed-form comparison, and we average over five shuffled answer-order seeds.

\begin{figure*}[t]
    \centering
    \includegraphics[width=0.86\textwidth]{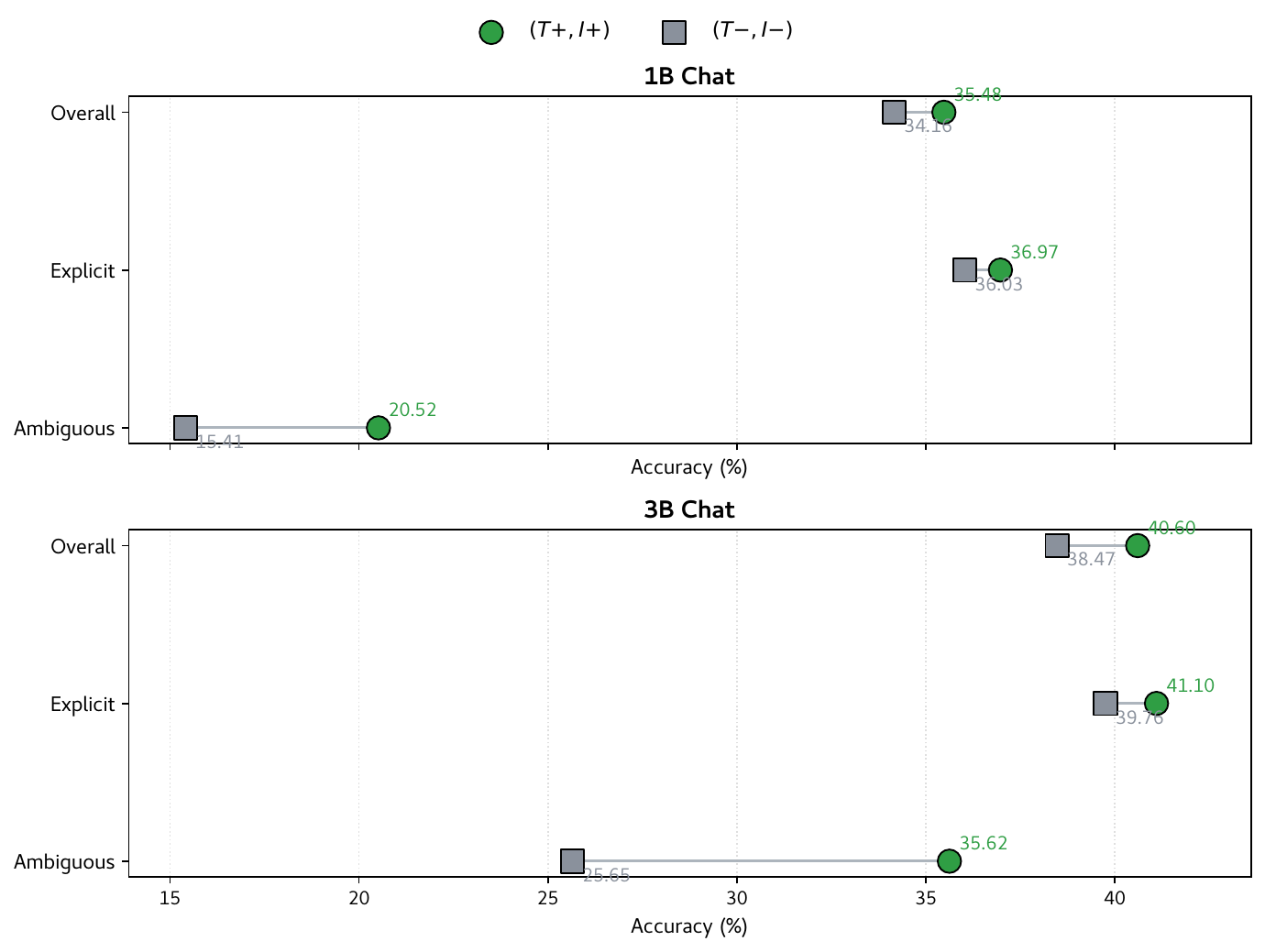}
    \caption{\textbf{\textsc{LocalNewsQA} after LoRA post-training.} Results are averaged over five shuffled option-order seeds.}
    \label{fig:qa-appendix}
\end{figure*}

Under this post-training view, the same directional pattern holds at both scales on the final gold benchmark. Figure~\ref{fig:qa-appendix} reports the full overall, explicit, and ambiguous accuracy matrix averaged over five answer-order seeds. The SFT results preserve the desired orderings, namely $(\Tplus, \Iplus)>(\Tminus, \Iminus)$ within each scale and $3$B$>1$B under full metadata across the three \textsc{LocalNewsQA} slices.

The same ordering also holds on the paired locale-switch metric after LoRA post-training. Full-metadata models improve both exact-pair success and margin-switch success over their metadata-free controls, with the larger chat model showing the stronger paired-switching rates. The appendix post-training comparison therefore preserves the same qualitative story on both one-sided target accuracy and paired two-sided locale switching.

\paragraph{Additional external \textsc{LocalNewsQA} baselines.}
We report the external instruct-model \textsc{LocalNewsQA} table alongside the other transfer evaluations later in this appendix so that all non-\textsc{MAPLE} reference baselines, including the frozen \textsc{LLaMA-3.2-1B} reference, appear in one place.

\section{Prompts}
\label{sec:prompts}
\begin{figure}[t]
    \centering
    \begin{tcolorbox}[
        enhanced,
        colback=teal!20,
        colframe=teal!90!black,
        coltext=teal,
        fontupper=\ttfamily\footnotesize,
        title=\centering \texttt{LocalNewsQA Instruction-Tuning Prompt},
        fonttitle=\bfseries\color{teal},
        coltitle=white,
        colbacktitle=teal!50,
        width=0.48\textwidth,
        boxrule=0.5mm,
        arc=3mm,
        top=1mm, bottom=1mm,
        left=1mm, right=1mm,
        boxsep=1.5mm
    ]
\textbf{SYSTEM}\quad Below is an instruction that describes a task, paired with an input that provides further context. Write a response that appropriately completes the request.\\[0.6em]
\textbf{USER}\\
URL = \{\{BASE\_URL\}\}/\{\{country\}\}\\
COUNTRY = \{\{country\}\}\\
CONTINENT = \{\{continent\}\}\\[0.4em]
TITLE = Locale-grounded QA about \{\{country\}\}\\[0.4em]
CONTENT\\
\#\#\# Instruction\\
\{\{example[instruction]\}\}\\[0.3em]
\#\#\# Input\\
\{\{example[input]\}\}\\[0.5em]
\textbf{ASSISTANT}\quad \{\{example[output]\}\}
    \end{tcolorbox}
    \caption{\textbf{Instruction-tuning prompt.} Prompt used for the \textsc{LocalNewsQA} post-training protocol.}
    \label{fig:sft_prompt}
\end{figure}

\begin{figure}[t]
    \centering
    \begin{tcolorbox}[
        enhanced,
        colback=teal!20,
        colframe=teal!90!black,
        coltext=teal,
        fontupper=\ttfamily\footnotesize,
        title=\centering \texttt{LocalNewsQA Evaluation Prompt},
        fonttitle=\bfseries\color{teal},
        coltitle=white,
        colbacktitle=teal!50,
        width=0.48\textwidth,
        boxrule=0.5mm,
        arc=3mm,
        top=1mm, bottom=1mm,
        left=1mm, right=1mm,
        boxsep=1.5mm
    ]
Below is an instruction that describes a task. Write a response that appropriately completes the request.\\[0.5em]
URL = \{\{BASE\_URL\}\}/\{\{country\_code\}\}\\
COUNTRY = \{\{country\_code\}\}\\
CONTINENT = \{\{Continent\}\}\\[0.5em]
TITLE = Locale-grounded QA about \{\{country\_code\}\}\\[0.5em]
CONTENT\\
Question = \{\{question\}\}\\[0.4em]
Options\\
A = \{\{option\_A\}\}\\
B = \{\{option\_B\}\}\\
C = \{\{option\_C\}\}\\
D = \{\{option\_D\}\}\\[0.4em]
Answer with the correct option.
    \end{tcolorbox}
    \caption{\textbf{LoRA evaluation prompt.} Prompt used for \textsc{LocalNewsQA} under the LoRA post-training protocol.}
    \label{fig:eval_prompt}
\end{figure}

\begin{figure*}[t]
    \centering
    \begin{tcolorbox}[
        enhanced,
        colback=teal!12,
        colframe=teal!90!black,
        coltext=black,
        title=\centering \texttt{LocalNewsQA Evaluation Prompt Examples},
        fonttitle=\bfseries\color{teal},
        coltitle=white,
        colbacktitle=teal!50,
        width=\textwidth,
        boxrule=0.5mm,
        arc=3mm,
        top=1mm, bottom=1mm,
        left=1.5mm, right=1.5mm,
        boxsep=1.0mm
    ]
    \begin{minipage}[t]{0.485\textwidth}
    \raggedright
    \textbf{\texttt{MAPLE 1B Evaluation Prompt}}\\[-0.1em]
    \textbf{BOS}\quad no tokenizer BOS token prepended before the prompt.\\[0.35em]
    {\ttfamily\scriptsize
    URL = www.globalfactcheck.org/lk\\
    COUNTRY = lk\\
    COUNTRY\_NAME = Sri Lanka\\
    CONTINENT = Asia\\[0.35em]
    TITLE = Facts about the country lk\\[0.35em]
    CONTENT\\
    Use the locale metadata above as grounding. This question is about Sri Lanka.\\[0.35em]
    Question = Which department is usually cited when census releases and official population statistics are reported?\\[0.35em]
    Options\\
    A = Statistics Department of the Central Bank\\
    B = Office of the Registrar General \& Census Commissioner\\
    C = Registrar General's Department\\
    D = Department of Census and Statistics\\[0.35em]
    Answer with the exact text of the correct option.\\[0.35em]
    Final answer
    }
    \end{minipage}\hfill
    \begin{minipage}[t]{0.485\textwidth}
    \raggedright
    \textbf{\texttt{MAPLE 3B Evaluation Prompt}}\\[-0.1em]
    \textbf{BOS}\quad prepend the tokenizer BOS token before the prompt.\\[0.35em]
    {\ttfamily\scriptsize
    URL = www.globalfactcheck.org/lk\\
    COUNTRY = Sri Lanka\\
    COUNTRY\_CODE = lk\\
    CONTINENT = Asia\\[0.35em]
    TITLE = Facts about Sri Lanka\\[0.35em]
    CONTENT\\
    Use the locale metadata above as grounding. Answer for Sri Lanka, not a different country.\\[0.35em]
    Question = Which department is usually cited when census releases and official population statistics are reported?\\[0.35em]
    Options\\
    - Statistics Department of the Central Bank\\
    - Office of the Registrar General \& Census Commissioner\\
    - Registrar General's Department\\
    - Department of Census and Statistics\\[0.35em]
    Final answer
    }
    \end{minipage}
    \end{tcolorbox}
    \caption{\textbf{\textsc{MAPLE} evaluation prompt examples.} Prompts used for the main \textsc{LocalNewsQA} evaluation. BOS handling is stated explicitly because it is not visible in the prompt text itself.}
    \label{fig:figure9_prompts}
\end{figure*}

\section{\textsc{LocalNewsQA} Construction and Validation}
\label{sec:localnewsqa-appendix}

\paragraph{Dataset construction}

\textsc{LocalNewsQA} is released as a two-split core benchmark for localized knowledge disambiguation in English news. The two splits are \textit{LocalNewsQA-Core-Explicit} and \textit{LocalNewsQA-Core-Ambiguous}. The final gold release contains $18{,}700$ items across $17$ countries and $4$ continents, with $17{,}000$ explicit items and $1{,}700$ ambiguous items. Each country contributes $1{,}000$ explicit rows and $100$ ambiguous rows. It covers $8$ topic families, \LocalNewsQATopicLegend. The release includes split definitions, target/contrast metadata, evaluation prompts, evidence fields, and source-validation fields so future work can compare methods under the same paired locale-switch protocol.

The main risk in this benchmark is not generating enough questions; it is admitting rows where the locale dependence, answer choice, or evidence relation is only approximate. Candidate generation was therefore recall-oriented, while release was precision-oriented. We used $816$ \textsc{GPT-5.4} batch requests over country, split, topic, and year shards to obtain $41{,}002$ candidates. This candidate pool was not treated as the benchmark. Exact/near duplicate removal, option-integrity checks, answer-consistency checks, ambiguity checks, low-information filters, and leakage filters first produced a larger retained pool of $35{,}874$ usable candidates. The reported benchmark is the stricter gold subset selected from that pool. Each explicit row needs target-side source support and a direct question-answer-evidence relation; each ambiguous row needs the same relation on both target and contrast sides, different supported answers, natural locale-omitting wording, and no surface cue that reveals the target country. Figure~\ref{fig:benchmark_prompt_dev} shows the explicit and ambiguous developer prompts used for candidate generation (temperature $0.6$, top-p $0.9$). Because the benchmark is hypothesis-targeted by design, we connect its downstream results to the controlled language-modeling studies, which rely on held-out perplexity measurements rather than question generation, and to the design-guaranteed negative-control floor in Table~\ref{tab:qa-factorial}.

\begin{figure*}[t]
    \centering
    \begin{minipage}[t]{0.485\textwidth}
    \begin{tcolorbox}[
        enhanced,
        colback=teal!20,
        colframe=teal!90!black,
        coltext=teal,
        fontupper=\ttfamily\scriptsize,
        title=\centering \texttt{LocalNewsQA Generation Prompt - EXPLICIT},
        fonttitle=\bfseries\color{teal},
        coltitle=white,
        colbacktitle=teal!50,
        width=\textwidth,
        boxrule=0.5mm,
        arc=3mm,
        top=1mm, bottom=1mm,
        left=1mm, right=1mm,
        boxsep=1.2mm
    ]
You are building \texttt{LocalNewsQA-Core-Explicit}, a benchmark for localized factual knowledge grounded in English news from 2010 to 2024. Return only a compact JSON object with an \texttt{items} array. Each item must be a unique, country-specific, non-ambiguous multiple-choice question whose correct answer is fully determined by the question text itself.\\[0.35em]
\textbf{Hard requirements}\\
- The question must be answerable from widely reported public knowledge plausibly learnable from English news from 2010 to 2024.\\
- Focus on the requested country and topic.\\
- Use exactly 4 options with 1 correct answer and 3 plausible distractors.\\
- Avoid trick questions and avoid questions whose answer changed repeatedly unless the year is stated clearly.\\
- Prefer precise named entities, offices, institutions, sports, holidays, infrastructure, or country-specific public facts.\\
- Do not ask about private individuals, offensive content, or unverifiable trivia.\\
- Questions must be mutually unique within the batch.\\[0.35em]
\textbf{Each item object must contain}\\
\texttt{question}, \texttt{options}, \texttt{correct\_answer}, \texttt{distractors}, \texttt{country}, \texttt{continent}, \texttt{topic}, \texttt{year}, \texttt{ambiguity\_flag} (\texttt{false}), and \texttt{evidence\_hint}.
    \end{tcolorbox}
    \end{minipage}\hfill
    \begin{minipage}[t]{0.485\textwidth}
    \begin{tcolorbox}[
        enhanced,
        colback=teal!20,
        colframe=teal!90!black,
        coltext=teal,
        fontupper=\ttfamily\scriptsize,
        title=\centering \texttt{LocalNewsQA Generation Prompt - AMBIGUOUS},
        fonttitle=\bfseries\color{teal},
        coltitle=white,
        colbacktitle=teal!50,
        width=\textwidth,
        boxrule=0.5mm,
        arc=3mm,
        top=1mm, bottom=1mm,
        left=1mm, right=1mm,
        boxsep=1.2mm
    ]
You are building \texttt{LocalNewsQA-Core-Ambiguous}, a benchmark for locale disambiguation grounded in English news from 2010 to 2024. Return only a compact JSON object with an \texttt{items} array. Each item must be an intentionally under-specified multiple-choice question whose correct answer depends on the target locale, and the same question text should plausibly admit a different correct answer under a different locale.\\[0.35em]
\textbf{Hard requirements}\\
- The question must be plausible for English news from 2010 to 2024.\\
- The question text must omit the country name and remain under-specified on purpose.\\
- The target answer must be correct for the requested target country.\\
- The contrast answer must be a realistic answer for the requested contrast country.\\
- Use exactly 4 options and include both the target answer and contrast answer in the options.\\
- Use exactly 1 correct answer for the target locale.\\
- Avoid artificial wording like ``in this country'' or ``given the metadata.''\\
- Prefer domains where locale matters, including institutions, offices, sports terms, public education terms, national bodies, holidays, transport conventions, or media/cultural references.\\
- Questions must be mutually unique within the batch.\\[0.35em]
\textbf{Each item object must contain}\\
\texttt{question}, \texttt{options}, \texttt{correct\_answer}, \texttt{distractors}, \texttt{country}, \texttt{continent}, \texttt{target\_country}, \texttt{contrast\_country}, \texttt{target\_answer}, \texttt{contrast\_answer}, \texttt{topic}, \texttt{year}, \texttt{ambiguity\_flag} (\texttt{true}), and \texttt{evidence\_hint}.
    \end{tcolorbox}
    \end{minipage}
    \caption{\textbf{\textsc{LocalNewsQA} generation prompts.} Prompts used to produce explicit and ambiguous candidate questions.}
    \label{fig:benchmark_prompt_dev}
    \vspace{0.4em}
\end{figure*}

\paragraph{Automatic source validation}
The retained pool is automatically validated row by row before the final gold release is selected. We build search queries from the question, answer, locale, and evidence hint; retrieve candidate evidence URLs, titles, and snippets; and then certify the support heuristically over the snippet and, when needed, the fetched page text. The certification rule is split-specific. Explicit rows require target-side certification only. Ambiguous rows require certification for both the target answer under the target locale and the paired contrast answer under the contrast locale. The gold pass then tightens this rule. Rows must have answer options that contain the gold answer, no duplicate normalized question, no high-confidence leakage cue, no unresolved evidence warning, and a direct relation between the question, answer, locale, and evidence. Ambiguous rows additionally require different target and contrast answers and both-sided relation support. The final split is then balanced by country. Table~\ref{tab:localnewsqa_validation_summary} summarizes the release counts after gold filtering.

\begin{table}[t]
\centering
\scriptsize
\setlength{\tabcolsep}{2.5pt}
\begin{tabular}{@{}L{0.25\linewidth}rL{0.34\linewidth}r@{}}
\toprule
Split & Rows & \makecell[c]{Gold validation\\rule} & \makecell[c]{Rows\\released} \\
\midrule
Explicit & $17{,}000$ & Target certification + direct QA/evidence relation & $17{,}000$ \\
Ambiguous & $1{,}700$ & Target + contrast certification and relation support & $1{,}700$ \\
\midrule
Final gold release & $18{,}700$ & Split-specific strict rule & $18{,}700$ \\
\bottomrule
\end{tabular}
\caption{Split-specific validation coverage for the final \textsc{LocalNewsQA} gold release. Explicit rows are balanced at $1{,}000$ per country and ambiguous rows at $100$ per country.}
\label{tab:localnewsqa_validation_summary}
\end{table}

\subsection*{Human Validation Protocol}

Automatic certification provides dataset-wide source grounding, but human validation remains necessary as an agreement audit over the same attached evidence. The human study asks whether annotators, when shown the question, answer options, locales, and retrieved sources, agree that the item is a valid benchmark example.

\paragraph{Study design}
We use a stratified $100$-item review sheet sampled from the final gold release, with $50$ explicit rows and $50$ ambiguous rows. The sample covers all $17$ countries and all $8$ topic families. The same $100$ rows are reviewed by three annotators so that agreement can be computed over identical items. The audit is not used to select the benchmark rows; it checks the already frozen gold split.

The audit records target-answer factuality for every row. Ambiguous rows additionally record contrast-answer factuality and whether the correct answer genuinely differs across the target and contrast locales. Each row receives a final accept/reject decision with one rejection reason when applicable.

\begin{table}[t]
\centering
\scriptsize
\setlength{\tabcolsep}{3pt}
\begin{tabular}{@{}lrrrr@{}}
\toprule
Sample slice & Rows & \makecell[c]{Majority\\accepted} & Accept rate & \makecell[c]{Wilson\\95\% CI} \\
\midrule
All audited rows & $100$ & $96$ & $96.0$ & $90.2$--$98.4$ \\
Explicit rows & $50$ & $50$ & $100.0$ & $92.9$--$100.0$ \\
Ambiguous rows & $50$ & $46$ & $92.0$ & $81.2$--$96.8$ \\
\bottomrule
\end{tabular}
\caption{Human-audit acceptance rates on a stratified sample from the final \textsc{LocalNewsQA} gold release. Confidence intervals are Wilson intervals for the majority accept rate.}
\label{tab:human-audit-acceptance}
\end{table}

\begin{table*}[t]
\centering
\scriptsize
\setlength{\tabcolsep}{3pt}
\begin{tabular}{@{}L{0.19\textwidth}ccccL{0.28\textwidth}@{}}
\toprule
Judgment & \makecell[c]{Unanimous\\agreement} & \makecell[c]{Avg. pairwise\\agreement} & Fleiss' $\kappa$ & \makecell[c]{Pairwise\\Cohen's $\kappa$} & \makecell[c]{Majority\\summary} \\
\midrule
Binary accept/reject & $97.0$ & $98.0$ & $0.759$ & $.852/.653/.790$ & $96$ accept, $4$ reject \\
Full rejection reason & $97.0$ & $97.7$ & $0.721$ & $.852/.656/.689$ & $96$ accept, $3$ weak-locale, $1$ tied reason \\
Target factuality & $98.0$ & $98.7$ & -- & -- & $100$ yes \\
Contrast factuality & $98.0$ & $98.7$ & $0.973$ & -- & $59$ yes, $40$ N/A, $1$ unsure \\
Locale dependence & $99.0$ & $99.3$ & $0.987$ & -- & $59$ yes, $37$ N/A, $3$ no, $1$ unsure \\
\bottomrule
\end{tabular}
\caption{Three-annotator agreement on the $100$-row human audit. Agreement columns are percentages. Pairwise Cohen's $\kappa$ values are reported for the two main decision labels. Fleiss' $\kappa$ is omitted when all majority labels fall into a single category, since chance-corrected agreement is not defined under zero category variation.}
\label{tab:human-audit-agreement}
\end{table*}

\begin{table*}[t]
\centering
\scriptsize
\setlength{\tabcolsep}{3pt}
\begin{tabular}{@{}L{0.16\textwidth}L{0.20\textwidth}L{0.42\textwidth}L{0.16\textwidth}@{}}
\toprule
ID & Locale pair & Question sketch & Majority decision \\
\midrule
\texttt{ambig\_0210} & Jamaica / United States & Prime minister during Office of the Contractor General governance stories & Weak locale difference \\
\texttt{ambig\_0623} & Sri Lanka / India & President during flagship infrastructure development coverage & Weak locale difference \\
\texttt{ambig\_0818} & Malaysia / United Kingdom & Land public transport regulator before agency restructuring & Weak locale difference \\
\texttt{ambig\_0280} & Jamaica / United States & Transport facility mentioned in local coverage about flights and redevelopment & Reject, reason tied \\
\bottomrule
\end{tabular}
\caption{Rows rejected by binary majority in the human audit. The first three rows have unanimous weak-locale rejections. The fourth row is a binary reject by majority, but annotators split on the exact reason, so the reason label is reported as tied.}
\label{tab:human-audit-rejected-rows}
\end{table*}

\begin{figure*}[t]
    \centering
    \begin{subfigure}[t]{0.49\textwidth}
        \centering
        \includegraphics[width=\textwidth]{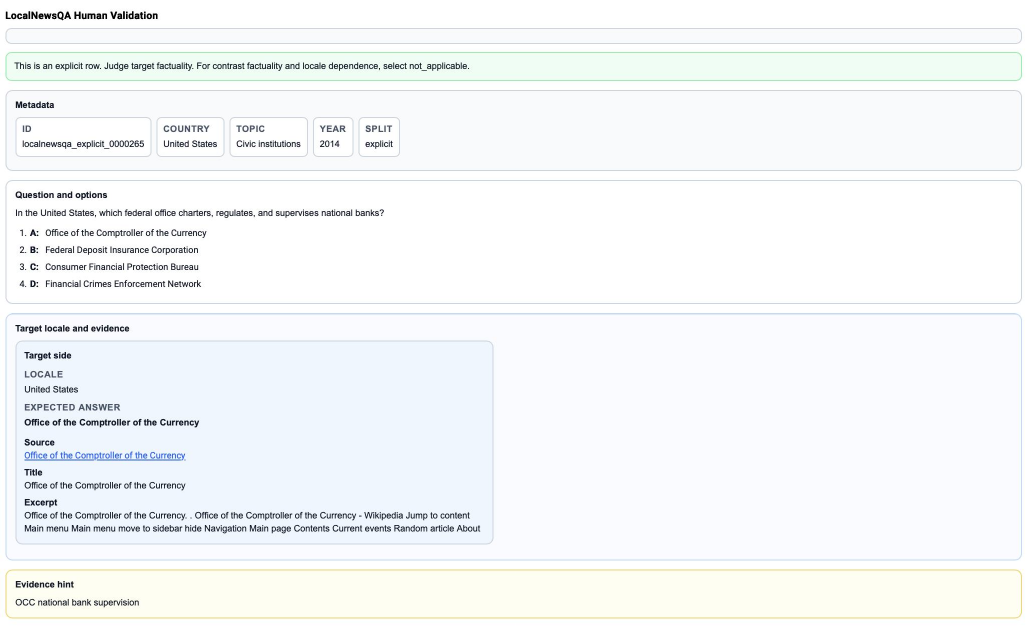}
        \caption{Explicit-row task view. Annotators see the question, answer options, target-locale evidence, and the explicit-row instruction banner.}
    \end{subfigure}\hfill
    \begin{subfigure}[t]{0.49\textwidth}
        \centering
        \includegraphics[width=\textwidth]{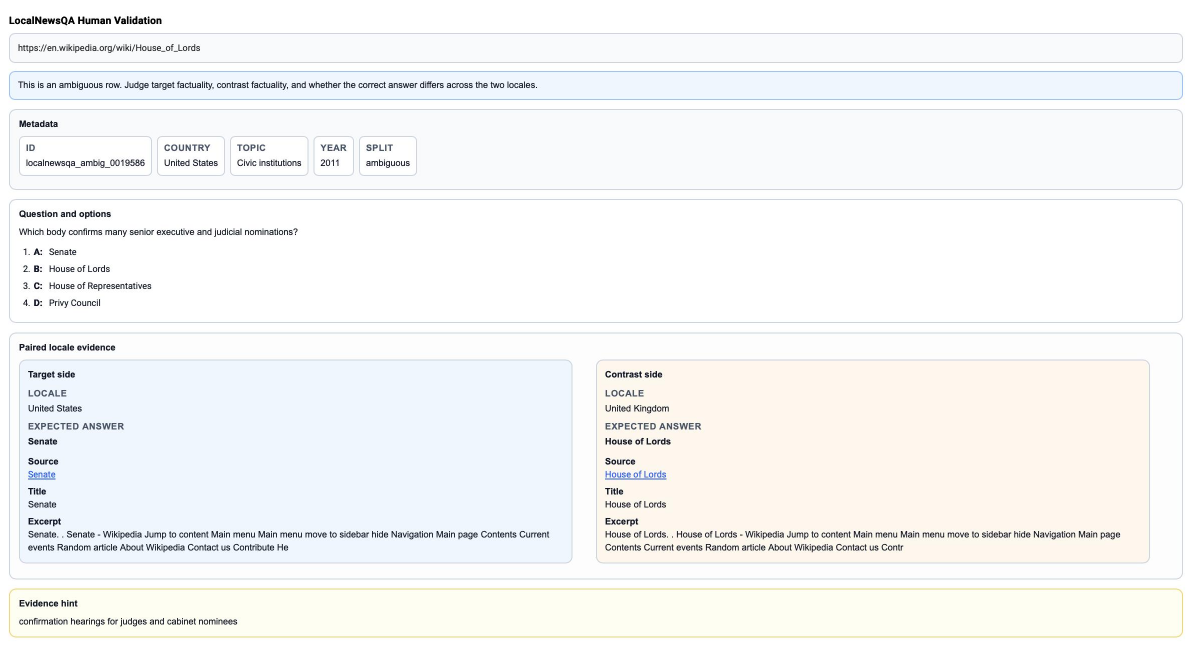}
        \caption{Ambiguous-row task view. Annotators see paired target and contrast evidence panels together with the locale-difference prompt.}
    \end{subfigure}

    \vspace{0.5em}

    \begin{subfigure}[t]{0.98\textwidth}
        \centering
        \includegraphics[width=\textwidth]{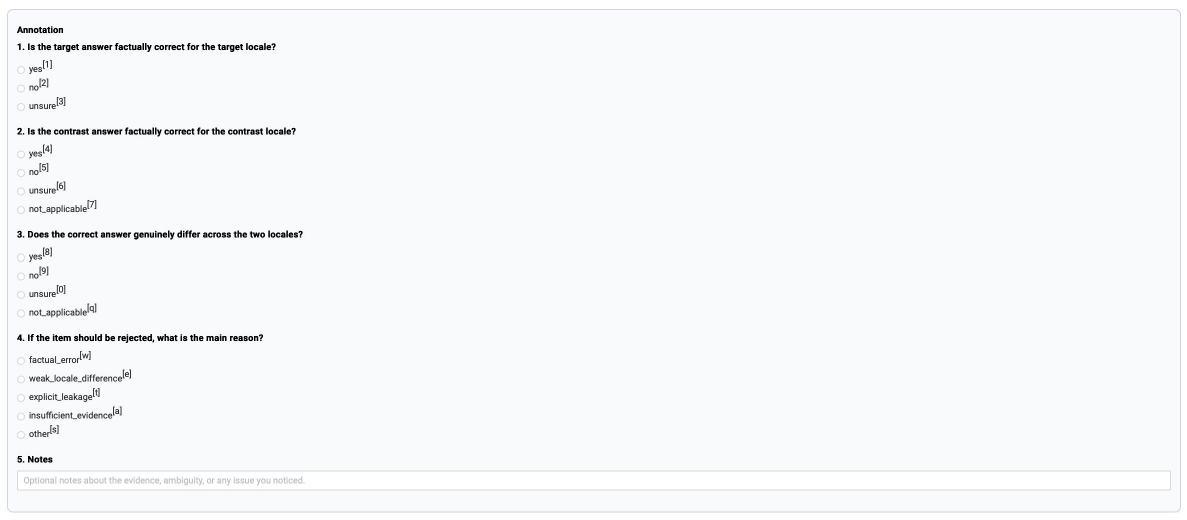}
        \caption{Judgment controls used to record target factuality, contrast factuality, locale dependence, rejection reason, and notes.}
    \end{subfigure}
    \caption{\textbf{Human-validation interface.} Label Studio task pages present explicit and ambiguous rows with their evidence context and use a shared judgment panel for annotation.}
    \label{fig:annotation-interface}
\end{figure*}

\paragraph{Annotation interface}
The annotation interface surfaces the question text, multiple-choice options, locale tags, evidence links, evidence snippets, and review fields in a single task view. The study materials are loaded from a CSV import in Label Studio, which presents the evidence panels and judgment controls together for each row. Figure~\ref{fig:annotation-interface} shows the interface used for the human-validation sheet.

\paragraph{Instructions for annotators}
Each annotator receives a short standardized protocol.
\begin{enumerate}
\item Read the question, options, locale fields, and attached evidence before making any judgment. Search beyond the attached evidence only if a displayed link is broken.
\item For \textbf{explicit} rows, answer one required question. \emph{Is the highlighted target answer correct for the displayed target locale?} Choose \texttt{yes}, \texttt{no}, or \texttt{unsure}.
\item For \textbf{ambiguous} rows, answer three required questions. \emph{Is the target answer correct for the target locale?} \emph{Is the contrast answer correct for the contrast locale?} \emph{Does the correct answer genuinely change across the two locales?} Each field is marked \texttt{yes}, \texttt{no}, or \texttt{unsure}.
\item If a row is rejected, optionally record one short reason, such as factual error, weak locale difference, explicit leakage, or insufficient evidence, together with a brief free-text note when needed.
\end{enumerate}

\paragraph{Participant-facing examples}
Figure~\ref{fig:annotation-examples} shows the two row types as participants see them. The explicit case requires one target-side judgment. The ambiguous case requires target-side, contrast-side, and locale-difference judgments.

\begin{figure*}[t]
    \centering
    \begin{subfigure}[t]{0.49\textwidth}
    \begin{tcolorbox}[
        appendixcard,
        colback=teal!12,
        colframe=teal!90!black,
        title=\centering \textbf{Explicit row shown to annotators},
        fonttitle=\bfseries,
        fontupper=\footnotesize,
        colbacktitle=teal!70!black,
        coltitle=white
    ]
    \setlength{\tabcolsep}{3pt}
	    \begin{tabular}{@{}L{0.22\linewidth}L{0.72\linewidth}@{}}
    \textbf{Target locale} & Hong Kong \\
    \textbf{Topic} & Education \\
    \textbf{Question} & In Hong Kong, which grades are referred to as primary school levels? \\
    \textbf{Options} & A. P1 to P6; B. S1 to S6; C. K1 to K3; D. Y7 to Y12 \\
    \textbf{Target answer} & P1 to P6 \\
    \textbf{Displayed evidence} & Po Leung Kuk Camoes Tan Siu Lin Primary School. \emph{``We have 5 classes for each class level from P1 to P6.''} \\
    \textbf{Recorded judgment} & Target correctness marked \texttt{yes} \\
    \end{tabular}
    \end{tcolorbox}
    \end{subfigure}
    \hfill
    \begin{subfigure}[t]{0.49\textwidth}
    \begin{tcolorbox}[
        appendixcard,
        colback=blue!6,
        colframe=blue!80!black,
        title=\centering \textbf{Ambiguous row shown to annotators},
        fonttitle=\bfseries,
        fontupper=\footnotesize,
        colbacktitle=blue!75!black,
        coltitle=white
    ]
    \setlength{\tabcolsep}{3pt}
	    \begin{tabular}{@{}L{0.22\linewidth}L{0.72\linewidth}@{}}
    \textbf{Target locale} & Canada \\
    \textbf{Contrast locale} & United States \\
    \textbf{Topic} & Education \\
    \textbf{Question} & A province expands full-day kindergarten for four- and five-year-olds. Which paired program name is most likely in the coverage? \\
    \textbf{Options} & A. JK-SK; B. Pre-K-K; C. TK-K; D. Nursery-Reception \\
    \textbf{Target answer} & JK-SK \\
    \textbf{Contrast answer} & Pre-K-K \\
    \textbf{Target evidence} & Kindergarten and Elementary, Avon Maitland District School Board. \emph{``Children who are 3 years old (JK) or 4 years old (SK) by December 31 are eligible to attend Full Day Kindergarten.''} \\
    \textbf{Contrast evidence} & Little Wildcats (Pre-K \& K), New Durham Elementary School. \emph{``Little Wildcats (Pre-K \& K).''} \\
    \textbf{Recorded judgments} & Target correctness marked \texttt{yes}; contrast correctness marked \texttt{yes}; locale difference marked \texttt{yes}. \\
    \end{tabular}
    \end{tcolorbox}
    \end{subfigure}
    \caption{\textbf{Representative participant views.} Explicit \textsc{LocalNewsQA} rows require one judgment. Ambiguous rows require three judgments tied to target correctness, contrast correctness, and whether the answer truly changes across locales.}
    \label{fig:annotation-examples}
\end{figure*}

\paragraph{Emissions estimate} Emissions from the API-based models used to construct the downstream data are negligible, fully offset by the cloud provider, and cost a total of $33$ USD. Total emissions from our on-premise GPU usage are estimated \citep{lacoste-2019-quantifying} at approximately $15$kgCO$_2$eq.

\renewcommand{\thesection}{A.\arabic{section}}
\section{Additional Analyses}
\label{sec:additional-analyses}

\begin{table}[t]
\centering
\scriptsize
\begin{tabular}{lrr}
\toprule
Signal & Count & Share (\%) \\
\midrule
Total rows & 15,633,804 & 100.0 \\
Any country-code signal & 6,680,266 & 42.7 \\
CC\,TLD present & 5,532,731 & 35.4 \\
CC\,TLD matches COUNTRY & 5,469,737 & 35.0 \\
Domain country-code token & 5,671,254 & 36.3 \\
Path country-code token & 1,323,947 & 8.5 \\
Any continent signal & 275,751 & 1.8 \\
Domain continent token & 38,496 & 0.25 \\
Path continent token & 238,381 & 1.5 \\
\bottomrule
\end{tabular}
\caption{URL-based geographic signals in the global training corpus (train split).}
\label{tab:url-signals}
\end{table}

\paragraph{Why does URL only outperform addition of other metadata?} Table~\ref{tab:url-signals} shows that URL text contains strong country-level signals in the global training corpus. About 42.7\% of URLs include an explicit country-code signal in the ccTLD (country-code top level domain), domain tokens, or path tokens, and ccTLDs alone appear in 35.4\% of URLs, matching the COUNTRY metadata in 35.0\% of cases. Path-level country codes are much less frequent (8.5\%), suggesting most country signal lives in the domain. In contrast, continent signals are rare (1.8\% overall), with domain-level continent tokens occurring in only 0.25\% of URLs. These findings indicate that URLs themselves carry substantial country-identifying information, which can plausibly drive locale-conditioned effects even without explicit metadata.

\paragraph{Mechanistic checks for URL provenance.}

The paper's central mechanistic comparison is URL provenance versus no metadata. Table~\ref{tab:url-signals} shows why this is the right contrast. URLs already carry substantial country-level signal in the format seen during pretraining, so the informative question is whether that signal leaves a detectable trace in the model's lexical expectations and hidden states relative to a metadata-free baseline.

\paragraph{Representation-level and lexical locale probes.}

The stronger mechanistic check is representation-level. On a held-out $1{,}000$-document global test set, we train linear probes on final-layer hidden states to recover continent and country labels. URL metadata separates cleanly from the no-metadata control. At the final checkpoint, a linear probe recovers continent from URL-conditioned hidden states at 91.5\% accuracy and country at 89.5\%, versus 67.5\% and 52.0\% from the no-metadata baseline (Table~\ref{tab:locale_probe_summary}). The URL-conditioned probe is already strong at $2$k steps and continues improving through $10$k, matching the paper's broader token-efficiency story.

As a smaller lexical check on the same $1$B pretrained family, we also constructed six short disambiguation prompts involving geographically ambiguous continuations such as \textit{football} $\rightarrow$ \textit{quarterback} vs.\ \textit{striker}, \textit{gas} vs.\ \textit{petrol}, and \textit{president} vs.\ \textit{prime minister}. For each prompt, we compare URL metadata against a no-metadata baseline. Table~\ref{tab:lexical_probe_summary} reports the mean symmetric KL divergence between the United States and United Kingdom next-token distributions, where higher means the model reacts more differently to the two locales, and the mean rank of the locale-appropriate continuation, where lower is better.

\begin{table}[t]
\centering
\scriptsize
\setlength{\tabcolsep}{5pt}
\resizebox{\columnwidth}{!}{%
\begin{tabular}{@{}lcc@{}}
\toprule
Prompt variant & Mean US to UK sym.\ KL $\uparrow$ & Mean expected rank $\downarrow$ \\
\midrule
URL metadata & \textbf{0.095} & \textbf{64.17} \\
No metadata & 0.000 & 200.96 \\
\bottomrule
\end{tabular}
}
\caption{Exploratory lexical locale-projection summary on six ambiguous continuations using the final $1$B pretrained checkpoints. No-metadata prompts collapse to identical distributions across locales, while URL metadata induces measurable locale sensitivity and gives much better rank to the locale-appropriate continuation.}
\label{tab:lexical_probe_summary}
\end{table}

\begin{table}[t]
\centering
\scriptsize
\setlength{\tabcolsep}{4pt}
\resizebox{\columnwidth}{!}{%
\begin{tabular}{@{}lcc@{}}
\toprule
Prompt variant & Continent probe acc.\ ($2$k $\rightarrow$ final) & Country probe acc.\ ($2$k $\rightarrow$ final) \\
\midrule
URL metadata & \textbf{89.5 $\rightarrow$ 91.5} & \textbf{88.0 $\rightarrow$ 89.5} \\
No metadata & 49.5 $\rightarrow$ 67.5 & 40.5 $\rightarrow$ 52.0 \\
\bottomrule
\end{tabular}
}
\caption{Linear-probe accuracy (\%) from final-layer hidden states on a held-out $1{,}000$-document global test set ($250$ per continent). We probe the representation at the last metadata token for URL-conditioned prompts and at the last title token for the no-metadata baseline. The informative comparison is URL versus no metadata.}
\label{tab:locale_probe_summary}
\end{table}

The lexical probe is intentionally small, but it points in the same direction. URL metadata induces locale-sensitive movement relative to the no-metadata control, which stays completely flat across locales on this synthetic next-token test, and it gives a substantially better average rank to the expected locale-appropriate continuation. The conservative mechanistic takeaway is that URL provenance produces substantially more locale-discriminative internal states than the metadata-free baseline while also shifting local lexical expectations in the expected direction.

\begin{figure*}[t]
    \centering
    \begin{subfigure}[t]{0.32\textwidth}
        \centering
        \includegraphics[width=\linewidth]{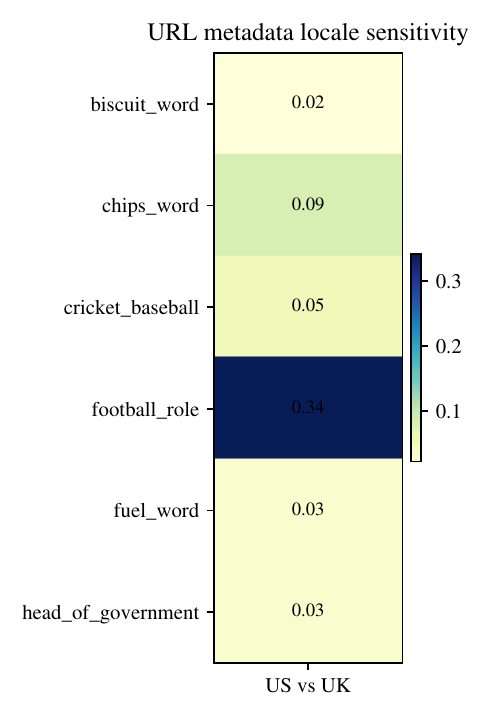}
        \caption{URL}
    \end{subfigure}
    \hfill
    \begin{subfigure}[t]{0.32\textwidth}
        \centering
        \includegraphics[width=\linewidth]{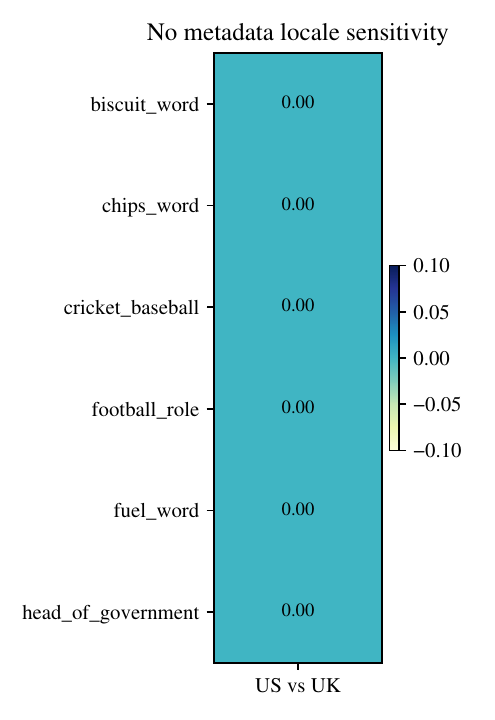}
        \caption{None}
    \end{subfigure}
    \caption{\textbf{Exploratory lexical locale-projection heatmaps.} Each row is one ambiguous lexical prompt, and each cell is the symmetric KL divergence between the United States and United Kingdom next-token distributions for that same prompt. Higher values mean the model reacts more differently to locale on the same context. The no-metadata model is effectively invariant, while URL metadata induces clear locale-sensitive movement.}
    \label{fig:lexical_probe_heatmaps}
\end{figure*}

\begin{figure}[t]
    \centering
    \begin{tcolorbox}[
        enhanced,
        colback=teal!20,
        colframe=teal!90!black,
        coltext=teal,
        fontupper=\ttfamily\footnotesize,
        title=\centering \texttt{URL Anatomy Example},
        fonttitle=\bfseries\color{teal},
        coltitle=white,
        colbacktitle=teal!50,
        width=\columnwidth,
        boxrule=0.5mm,
        arc=3mm,
        top=1mm, bottom=1mm,
        left=1mm, right=1mm,
        boxsep=2mm
    ]
\textbf{URL}\quad \texttt{www.bbc.co.uk}\\
\hspace*{1.3em}\texttt{news/world-europe-12345}\\[0.6em]
\textbf{ccTLD}\quad \texttt{uk}\\
\textbf{Domain}\quad \texttt{bbc.co.uk}\\
\textbf{Path}\quad \texttt{/news/world-europe-12345}
    \end{tcolorbox}
    \caption{\textbf{URL signal extraction example.} Example of ccTLD, domain, and path extraction from a URL.}
    \label{fig:url_anatomy}
\end{figure}

\paragraph{Cluster-level analysis of leave-one-out effects.}

To better understand why the \highlightRoundedBorder{white}{[NoAmerica]} model exhibits the smallest degradation relative to the full global model in Figure~\ref{fig:ablation-leaveone}, we perform a cluster-level analysis on the combined test set. We compute sentence representations by mean-pooling the final-layer hidden states produced by the \highlightRoundedBorder{white}{[NoAmerica]} model and apply $k$-means clustering with $K=4$, mirroring the four training continents. Using embeddings from the \highlightRoundedBorder{white}{[NoAmerica]} model lets us inspect how the global test set is organized when America-specific training data has been removed.

We first compare cluster assignments with continent labels. As Table~\ref{tab:cluster_continent_mix} shows, continent labels are broadly mixed across clusters rather than forming continent-specific partitions. This indicates substantial cross-continent overlap in the combined test set and helps explain why omitting America produces a smaller degradation than omitting some of the other continents because the remaining training data still covers a relatively broad portion of the latent structure of the evaluation corpus.

\begin{table}[t]
\centering
\scriptsize
\begin{tabular}{@{}lcccc@{}}
\toprule
Cluster (n) & Africa & America & Asia & Europe \\
\midrule
0 (321) & 132 (41.1\%) & 32 (10.0\%) & 125 (38.9\%) & 32 (10.0\%) \\
1 (237) & 33 (13.9\%) & 72 (30.4\%) & 62 (26.2\%) & 70 (29.5\%) \\
2 (99)  & 29 (29.3\%) & 12 (12.1\%) & 30 (30.3\%) & 28 (28.3\%) \\
3 (343) & 42 (12.2\%) & 146 (42.6\%) & 29 (8.5\%) & 126 (36.7\%) \\
\bottomrule
\end{tabular}
\caption{Continent distribution per cluster on the combined test set (NoAmerica embeddings).}
\label{tab:cluster_continent_mix}
\end{table}

\paragraph{External downstream test sets.}

Table~\ref{tab:qa-factorial} gives the four-setting \textsc{MAPLE} paired-switch comparison. The appendix tables below report complementary endpoint gains with paired bootstrap intervals for \textsc{LocalNewsQA} and the external benchmarks used in the transfer analysis. On the \textsc{LocalNewsQA} ambiguous split, the aggregate gains are clear and the paired metrics are positive, while granular country/topic target-accuracy buckets are reported as diagnostics rather than as a universal-positivity claim.

\begin{table*}[t]
\centering
\scriptsize
\setlength{\tabcolsep}{2.0pt}
\begin{tabular*}{\textwidth}{@{\extracolsep{\fill}}lccccc@{}}
\toprule
Model & \makecell[c]{Overall\\[-1pt]$\Delta$ acc.} & \makecell[c]{Ambig.\\[-1pt]$\Delta$ acc.} & \makecell[c]{Explicit\\[-1pt]$\Delta$ acc.} & \makecell[c]{Exact pair\\[-1pt]$\Delta$} & \makecell[c]{Margin switch\\[-1pt]$\Delta$} \\
\midrule
\multicolumn{6}{@{}l}{\textit{\providerMAPLE{}~MAPLE family}\hfill {\scriptsize \tableprimaryplus{$(\Tplus,\Iplus)$}$-$\tableprimaryminus{$(\Tminus,\Iminus)$}}} \\
MAPLE 1B Base & \cigainposstrong{+1.4}{+0.9}{+1.9} & \cigainposstrong{+6.7}{+5.1}{+8.4} & \cigainposstrong{+0.9}{+0.3}{+1.4} & \cigainposstrong{+1.9}{+1.2}{+2.5} & \cigainposstrong{+8.2}{+6.8}{+9.6} \\
MAPLE 1B Chat & \cigainposstrong{+1.2}{+0.6}{+1.9} & \cigainposstrong{+4.4}{+2.4}{+6.5} & \cigainposstrong{+0.9}{+0.2}{+1.6} & \cigainposstrong{+2.0}{+1.4}{+2.7} & \cigainposstrong{+6.5}{+5.4}{+7.7} \\
MAPLE 3B Base & \cigainposstrong{+2.8}{+2.0}{+3.6} & \cigainposstrong{+9.2}{+6.5}{+11.8} & \cigainposstrong{+2.1}{+1.2}{+3.0} & \cigainposstrong{+8.8}{+7.4}{+10.1} & \cigainposstrong{+25.1}{+23.1}{+27.2} \\
MAPLE 3B Chat & \cigainposstrong{+2.2}{+1.4}{+3.0} & \cigainposstrong{+10.1}{+7.5}{+12.8} & \cigainposstrong{+1.4}{+0.5}{+2.2} & \cigainposstrong{+5.1}{+4.1}{+6.2} & \cigainposstrong{+13.6}{+12.1}{+15.2} \\
\midrule
\multicolumn{6}{@{}l}{\textit{\providerMeta{}~LLaMA family}\hfill {\scriptsize \tableprimaryplus{$\Iplus$}$-$\tableprimaryminus{$\Iminus$}}} \\
LLaMA-3.2-1B Base & \cigainnegstrong{-1.2}{-1.6}{-0.8} & \cigainneg{-0.1}{-1.5}{+1.3} & \cigainnegstrong{-1.3}{-1.8}{-0.9} & \cigainposstrong{+1.4}{+0.8}{+1.9} & \cigainposstrong{+7.4}{+6.2}{+8.6} \\
LLaMA-3.2-1B Inst. & \cigainnegstrong{-2.3}{-2.7}{-1.8} & \cigainpos{+0.4}{-1.1}{+1.9} & \cigainnegstrong{-2.5}{-3.0}{-2.1} & \cigainposstrong{+3.1}{+2.3}{+3.9} & \cigainposstrong{+9.2}{+7.8}{+10.5} \\
LLaMA-3.2-3B Base & \cigainnegstrong{-1.0}{-1.4}{-0.5} & \cigainposstrong{+5.2}{+3.7}{+6.7} & \cigainnegstrong{-1.6}{-2.1}{-1.2} & \cigainposstrong{+5.0}{+4.0}{+6.1} & \cigainposstrong{+16.2}{+14.5}{+17.9} \\
LLaMA-3.2-3B Inst. & \cigainnegstrong{-8.0}{-8.6}{-7.5} & \cigainnegstrong{-2.6}{-4.5}{-0.9} & \cigainnegstrong{-8.5}{-9.1}{-8.0} & \cigainposstrong{+3.6}{+2.8}{+4.5} & \cigainposstrong{+10.6}{+9.2}{+12.1} \\
\midrule
\multicolumn{6}{@{}l}{\textit{\providerQwen{}~Qwen2.5 family}\hfill {\scriptsize \tableprimaryplus{$\Iplus$}$-$\tableprimaryminus{$\Iminus$}}} \\
Qwen2.5-1.5B Base & \cigainnegstrong{-0.7}{-1.1}{-0.3} & \cigainposstrong{+1.4}{+0.2}{+2.6} & \cigainnegstrong{-0.9}{-1.3}{-0.4} & \cigainposstrong{+1.5}{+0.9}{+2.1} & \cigainposstrong{+5.4}{+4.4}{+6.5} \\
Qwen2.5-1.5B Inst. & \cigainnegstrong{-1.8}{-2.3}{-1.4} & \cigainposstrong{+8.5}{+6.8}{+10.4} & \cigainnegstrong{-2.9}{-3.3}{-2.4} & \cigainposstrong{+7.7}{+6.5}{+8.9} & \cigainposstrong{+20.4}{+18.5}{+22.4} \\
Qwen2.5-3B Base & \cigainnegstrong{-2.6}{-3.0}{-2.1} & \cigainposstrong{+8.8}{+7.1}{+10.6} & \cigainnegstrong{-3.7}{-4.2}{-3.3} & \cigainposstrong{+10.5}{+9.1}{+11.9} & \cigainposstrong{+25.3}{+23.3}{+27.3} \\
Qwen2.5-3B Inst. & \cigainposstrong{+0.6}{+0.2}{+1.0} & \cigainposstrong{+8.7}{+7.1}{+10.4} & \cigainneg{-0.2}{-0.6}{+0.2} & \cigainposstrong{+8.2}{+6.9}{+9.6} & \cigainposstrong{+25.2}{+23.2}{+27.4} \\
\midrule
\multicolumn{6}{@{}l}{\textit{\providerQwen{}~Qwen3.5 family}\hfill {\scriptsize \tableprimaryplus{$\Iplus$}$-$\tableprimaryminus{$\Iminus$}}} \\
Qwen3.5-0.8B Base & \cigainnegstrong{-0.7}{-1.1}{-0.4} & \cigainposstrong{+2.5}{+1.1}{+3.8} & \cigainnegstrong{-1.1}{-1.4}{-0.7} & \cigainposstrong{+2.7}{+1.9}{+3.5} & \cigainposstrong{+8.5}{+7.1}{+9.8} \\
Qwen3.5-0.8B Chat & \cigainnegstrong{-1.1}{-1.4}{-0.7} & \cigainposstrong{+3.8}{+2.4}{+5.1} & \cigainnegstrong{-1.6}{-1.9}{-1.2} & \cigainposstrong{+2.2}{+1.6}{+3.0} & \cigainposstrong{+8.3}{+6.9}{+9.6} \\
Qwen3.5-2B Base & \cigainposstrong{+1.0}{+0.7}{+1.4} & \cigainposstrong{+9.6}{+8.1}{+11.2} & \cigainpos{+0.2}{-0.1}{+0.5} & \cigainposstrong{+9.2}{+7.8}{+10.5} & \cigainposstrong{+22.4}{+20.4}{+24.4} \\
Qwen3.5-2B Chat & \cigainposstrong{+2.7}{+2.2}{+3.1} & \cigainposstrong{+11.2}{+9.4}{+13.0} & \cigainposstrong{+1.8}{+1.4}{+2.2} & \cigainposstrong{+8.5}{+7.2}{+9.9} & \cigainposstrong{+20.4}{+18.6}{+22.4} \\
Qwen3.5-4B Base & \cigainpos{+0.1}{-0.1}{+0.3} & \cigainposstrong{+1.5}{+0.7}{+2.2} & \cigainzero{0.0}{-0.3}{+0.2} & \cigainposstrong{+1.2}{+0.7}{+1.7} & \cigainposstrong{+3.6}{+2.8}{+4.5} \\
Qwen3.5-4B Chat & \cigainpos{+0.2}{-0.1}{+0.4} & \cigainposstrong{+1.3}{+0.5}{+2.2} & \cigainpos{+0.1}{-0.2}{+0.3} & \cigainposstrong{+0.8}{+0.4}{+1.3} & \cigainposstrong{+3.1}{+2.4}{+4.1} \\
\midrule
\multicolumn{6}{@{}l}{\textit{\providerGoogle{}~Gemma-4 family}\hfill {\scriptsize \tableprimaryplus{$\Iplus$}$-$\tableprimaryminus{$\Iminus$}}} \\
Gemma-4-E2B Base & \cigainposstrong{+2.0}{+1.5}{+2.5} & \cigainneg{-0.1}{-1.8}{+1.6} & \cigainposstrong{+2.2}{+1.7}{+2.8} & \cigainposstrong{+0.5}{+0.2}{+0.9} & \cigainposstrong{+2.3}{+1.6}{+3.0} \\
Gemma-4-E2B-it & \cigainposstrong{+1.0}{+0.7}{+1.3} & \cigainposstrong{+2.8}{+1.5}{+4.1} & \cigainposstrong{+0.8}{+0.5}{+1.2} & \cigainposstrong{+3.5}{+2.6}{+4.4} & \cigainposstrong{+9.8}{+8.5}{+11.4} \\
Gemma-4-E4B Base & \cigainpos{+0.5}{0.0}{+1.0} & \cigainneg{-0.9}{-2.5}{+0.8} & \cigainposstrong{+0.6}{+0.1}{+1.2} & \cigainposstrong{+0.6}{+0.1}{+1.1} & \cigainposstrong{+2.3}{+1.4}{+3.2} \\
Gemma-4-E4B-it & \cigainposstrong{+2.7}{+2.3}{+3.0} & \cigainposstrong{+7.1}{+5.6}{+8.6} & \cigainposstrong{+2.2}{+1.9}{+2.6} & \cigainposstrong{+8.8}{+7.5}{+10.1} & \cigainposstrong{+23.2}{+21.2}{+25.3} \\
\midrule
\multicolumn{6}{@{}l}{\textit{\providerMistral{}~Ministral family}\hfill {\scriptsize \tableprimaryplus{$\Iplus$}$-$\tableprimaryminus{$\Iminus$}}} \\
Ministral-3-3B Base & \cigainposstrong{+0.7}{+0.3}{+1.1} & \cigainposstrong{+4.8}{+3.5}{+6.1} & \cigainpos{+0.3}{-0.2}{+0.7} & \cigainposstrong{+4.5}{+3.5}{+5.5} & \cigainposstrong{+16.2}{+14.5}{+18.1} \\
Ministral-3-3B Inst. & \cigainposstrong{+0.9}{+0.6}{+1.3} & \cigainposstrong{+6.4}{+4.9}{+7.9} & \cigainposstrong{+0.4}{+0.1}{+0.7} & \cigainposstrong{+5.6}{+4.6}{+6.9} & \cigainposstrong{+18.5}{+16.7}{+20.4} \\
\bottomrule
\end{tabular*}

\caption{\textsc{LocalNewsQA} metadata gains with paired bootstrap $95\%$ confidence intervals. Each completed cell reports the point estimate on top and the confidence interval below. Accuracy columns report percentage-point gains on target-locale questions; exact-pair and margin-switch columns report percentage-point gains on ambiguous target/contrast pairs. \textsc{MAPLE} rows compare \tableprimaryplus{$(\Tplus, \Iplus)$}$-$\tableprimaryminus{$(\Tminus, \Iminus)$}, while non-\textsc{MAPLE} rows compare \tableprimaryplus{$\Iplus$}$-$\tableprimaryminus{$\Iminus$}. Positive values favor metadata conditioning or metadata-formatted inference; dark bold point estimates have intervals that exclude zero, while lighter entries overlap zero.}
\label{tab:locale_switch_pairs}
\label{tab:localnewsqa_model_gain_summary}
\end{table*}

\begin{table*}[t]
\centering
\scriptsize
\setlength{\tabcolsep}{2.0pt}
\begin{tabular*}{\textwidth}{@{\extracolsep{\fill}}lcccccc@{}}
\toprule
Model & \makecell[c]{GeoMLaMA\\[-1pt]$\Delta$ acc.} & \makecell[c]{GOQA\\[-1pt]$\Delta$ acc.} & \makecell[c]{MMLU-CS\\[-1pt]$\Delta$ acc.} & \makecell[c]{NormAD\\[-1pt]$\Delta$ acc.} & \makecell[c]{BLEnD\\[-1pt]$\Delta$ acc.} & \makecell[c]{WVB\\[-1pt]EMD red.} \\
\midrule
\multicolumn{7}{@{}l}{\textit{\providerMAPLE{}~MAPLE family}\hfill {\scriptsize \tableprimaryplus{$(\Tplus,\Iplus)$}$-$\tableprimaryminus{$(\Tminus,\Iminus)$}}} \\
MAPLE 1B Base & \cigainpos{+2.7}{-8.0}{+13.3} & \cigainposstrong{+1.3}{+0.4}{+2.1} & \cigainpos{+0.1}{-2.4}{+2.7} & \cigainposstrong{+1.4}{+0.3}{+2.6} & \cigainposstrong{+9.4}{+3.1}{+15.3} & \cigainposstrong{+0.03}{+0.01}{+0.05} \\
MAPLE 1B Chat & \cigainpos{+4.0}{-6.0}{+13.3} & \cigainposstrong{+2.7}{+1.8}{+3.6} & \cigainpos{+0.3}{-0.6}{+1.1} & \cigainposstrong{+3.8}{+0.9}{+6.4} & \cigainpos{+1.0}{-2.0}{+4.1} & \cigainposstrong{+0.01}{+0.01}{+0.02} \\
MAPLE 3B Base & \cigainpos{+5.3}{-4.7}{+14.7} & \cigainposstrong{+8.4}{+7.3}{+9.6} & \cigainposstrong{+3.9}{+0.5}{+7.3} & \cigainposstrong{+5.1}{+2.7}{+7.4} & \cigainposstrong{+13.5}{+7.1}{+19.6} & \cigainzero{0.00}{0.00}{+0.01} \\
MAPLE 3B Chat & \cigainpos{+4.7}{-4.7}{+13.3} & \cigainposstrong{+3.4}{+2.5}{+4.3} & \cigainpos{+0.7}{-0.1}{+1.5} & \cigainposstrong{+2.4}{+0.2}{+4.6} & \cigainposstrong{+6.1}{+0.8}{+11.7} & \cigainposstrong{+0.02}{0.00}{+0.04} \\
\midrule
\multicolumn{7}{@{}l}{\textit{\providerMeta{}~LLaMA family}\hfill {\scriptsize \tableprimaryplus{$\Iplus$}$-$\tableprimaryminus{$\Iminus$}}} \\
LLaMA-3.2-1B Base & \cigainnegstrong{-2.7}{-5.3}{-0.7} & \cigainneg{-0.1}{-0.5}{+0.3} & \cigainposstrong{+2.0}{+0.5}{+3.5} & \cigainpos{+1.4}{-0.5}{+3.5} & \cigainzero{0.0}{-2.0}{+2.0} & \cigainpos{+0.02}{-0.02}{+0.05} \\
LLaMA-3.2-1B Inst. & \cigainpos{+5.3}{-2.7}{+12.7} & \cigainneg{-0.2}{-0.6}{+0.1} & \cigainneg{-1.0}{-2.5}{+0.5} & \cigainzero{0.0}{0.0}{0.0} & \cigainneg{-1.3}{-4.1}{+1.3} & \cigainposstrong{+0.03}{+0.02}{+0.04} \\
LLaMA-3.2-3B Base & \cigainneg{-5.3}{-14.0}{+2.7} & \cigainposstrong{+0.4}{+0.1}{+0.7} & \cigainpos{+0.8}{-0.5}{+2.0} & \cigainpos{+1.9}{-0.1}{+3.9} & \cigainpos{+1.5}{-0.3}{+3.3} & \cigainnegstrong{-0.03}{-0.04}{-0.02} \\
LLaMA-3.2-3B Inst. & \cigainneg{-4.7}{-12.7}{+4.0} & \cigainnegstrong{-1.0}{-1.4}{-0.6} & \cigainnegstrong{-2.9}{-4.8}{-1.0} & \cigainneg{-0.5}{-1.6}{+0.5} & \cigainnegstrong{-9.2}{-13.0}{-5.9} & \cigainnegstrong{-0.10}{-0.13}{-0.07} \\
\midrule
\multicolumn{7}{@{}l}{\textit{\providerQwen{}~Qwen2.5 family}\hfill {\scriptsize \tableprimaryplus{$\Iplus$}$-$\tableprimaryminus{$\Iminus$}}} \\
Qwen2.5-1.5B Base & \cigainnegstrong{-10.7}{-16.0}{-5.3} & \cigainnegstrong{-2.1}{-2.7}{-1.5} & \cigainpos{+0.8}{-1.0}{+2.4} & \cigainposstrong{+5.0}{+1.7}{+8.1} & \cigainneg{-0.8}{-2.8}{+1.3} & \cigainpos{+0.01}{0.00}{+0.01} \\
Qwen2.5-1.5B Inst. & \cigainnegstrong{-8.7}{-14.7}{-2.7} & \cigainnegstrong{-0.8}{-1.2}{-0.4} & \cigainneg{-0.3}{-1.9}{+1.4} & \cigainnegstrong{-2.7}{-4.1}{-1.4} & \cigainneg{-1.5}{-4.3}{+1.3} & \cigainnegstrong{-0.02}{-0.04}{-0.01} \\
Qwen2.5-3B Base & \cigainzero{0.0}{-5.3}{+6.0} & \cigainnegstrong{-0.4}{-0.8}{-0.1} & \cigainneg{-0.3}{-2.0}{+1.5} & \cigainposstrong{+9.6}{+7.8}{+11.4} & \cigainneg{-1.8}{-4.1}{+0.5} & \cigainnegstrong{-0.02}{-0.04}{-0.01} \\
Qwen2.5-3B Inst. & \cigainneg{-2.7}{-9.3}{+4.0} & \cigainneg{-0.2}{-0.7}{+0.2} & \cigainzero{0.0}{-1.8}{+1.8} & \cigainneg{-0.1}{-1.1}{+1.0} & \cigainpos{+1.3}{-1.5}{+4.1} & \cigainpos{+0.01}{0.00}{+0.02} \\
\midrule
\multicolumn{7}{@{}l}{\textit{\providerQwen{}~Qwen3.5 family}\hfill {\scriptsize \tableprimaryplus{$\Iplus$}$-$\tableprimaryminus{$\Iminus$}}} \\
Qwen3.5-0.8B Base & \cigainneg{-1.3}{-5.3}{+2.7} & \cigainzero{0.0}{-0.4}{+0.3} & \cigainpos{+0.8}{-0.8}{+2.3} & \cigainneg{-0.5}{-1.4}{+0.5} & \cigainneg{-0.8}{-2.8}{+1.3} & \cigainzero{0.00}{-0.01}{+0.01} \\
Qwen3.5-0.8B Chat & \cigainneg{-2.7}{-6.7}{+0.7} & \cigainneg{-0.2}{-0.6}{+0.1} & \cigainpos{+1.0}{-0.1}{+2.3} & \cigainnegstrong{-2.1}{-3.1}{-1.3} & \cigainneg{-2.0}{-4.3}{+0.3} & \cigainposstrong{+0.04}{+0.02}{+0.06} \\
Qwen3.5-2B Base & \cigainneg{-3.3}{-6.7}{0.0} & \cigainpos{+0.2}{-0.1}{+0.6} & \cigainpos{+0.6}{-0.8}{+2.1} & \cigainnegstrong{-2.5}{-3.6}{-1.5} & \cigainpos{+1.0}{-1.3}{+3.6} & \cigainpos{+0.01}{0.00}{+0.02} \\
Qwen3.5-2B Chat & \cigainpos{+1.3}{-3.3}{+6.7} & \cigainzero{0.0}{-0.4}{+0.4} & \cigainpos{+0.8}{-0.9}{+2.4} & \cigainnegstrong{-4.0}{-4.9}{-3.1} & \cigainpos{+0.3}{-2.3}{+2.5} & \cigainneg{-0.01}{-0.03}{+0.01} \\
Qwen3.5-4B Base & \cigainpos{+2.0}{-4.7}{+8.7} & \cigainneg{-0.2}{-0.5}{0.0} & \cigainpos{+0.5}{-0.5}{+1.5} & \cigainposstrong{+0.4}{+0.2}{+0.7} & \cigainpos{+0.5}{0.0}{+1.3} & \cigainzero{0.00}{0.00}{0.00} \\
Qwen3.5-4B Chat & \cigainnegstrong{-3.3}{-6.7}{-0.7} & \cigainneg{-0.2}{-0.5}{+0.1} & \cigainneg{-0.4}{-1.5}{+0.6} & \cigainzero{0.0}{0.0}{0.0} & \cigainnegstrong{-1.5}{-2.8}{-0.5} & \cigainzero{0.00}{0.00}{0.00} \\
\midrule
\multicolumn{7}{@{}l}{\textit{\providerGoogle{}~Gemma-4 family}\hfill {\scriptsize \tableprimaryplus{$\Iplus$}$-$\tableprimaryminus{$\Iminus$}}} \\
Gemma-4-E2B Base & \cigainpos{+2.7}{-1.3}{+6.7} & \cigainpos{+0.3}{-0.1}{+0.6} & \cigainpos{+0.3}{-1.3}{+1.8} & \cigainposstrong{+6.1}{+4.7}{+7.5} & \cigainneg{-0.3}{-2.0}{+1.5} & \cigainzero{0.00}{0.00}{0.00} \\
Gemma-4-E2B-it & \cigainpos{+2.0}{-6.0}{+10.0} & \cigainneg{-0.1}{-0.4}{+0.3} & \cigainpos{+0.5}{-1.5}{+2.5} & \cigainpos{+1.1}{0.0}{+2.2} & \cigainneg{-0.5}{-2.8}{+1.8} & \cigainposstrong{+0.03}{+0.01}{+0.05} \\
Gemma-4-E4B Base & \cigainpos{+5.3}{-2.7}{+12.7} & \cigainnegstrong{-0.6}{-0.9}{-0.2} & \cigainposstrong{+2.9}{+1.4}{+4.5} & \cigainposstrong{+1.7}{+0.9}{+2.5} & \cigainposstrong{+5.1}{+2.3}{+8.1} & \cigainzero{0.00}{0.00}{+0.01} \\
Gemma-4-E4B-it & \cigainpos{+1.3}{-3.3}{+6.0} & \cigainnegstrong{-1.0}{-1.4}{-0.6} & \cigainposstrong{+2.7}{+0.8}{+4.4} & \cigainposstrong{+2.5}{+1.2}{+3.8} & \cigainneg{-1.5}{-3.8}{+0.8} & \cigainneg{-0.01}{-0.02}{+0.01} \\
\midrule
\multicolumn{7}{@{}l}{\textit{\providerMistral{}~Ministral family}\hfill {\scriptsize \tableprimaryplus{$\Iplus$}$-$\tableprimaryminus{$\Iminus$}}} \\
Ministral-3-3B Base & \cigainneg{-3.3}{-8.7}{+1.3} & \cigainnegstrong{-0.3}{-0.6}{0.0} & \cigainzero{0.0}{-1.3}{+1.3} & \cigainnegstrong{-2.1}{-3.1}{-1.0} & \cigainneg{-0.5}{-3.1}{+2.0} & \cigainnegstrong{-0.04}{-0.06}{-0.01} \\
Ministral-3-3B Inst. & \cigainnegstrong{-12.7}{-18.7}{-7.3} & \cigainpos{+0.1}{-0.3}{+0.4} & \cigainnegstrong{-2.7}{-4.3}{-1.1} & \cigainnegstrong{-4.2}{-5.1}{-3.3} & \cigainneg{-0.8}{-2.5}{+1.0} & \cigainnegstrong{-0.02}{-0.03}{0.00} \\
\bottomrule
\end{tabular*}

\caption{External benchmark metadata gains with paired bootstrap $95\%$ confidence intervals, grouped by model family. Each completed cell reports the point estimate on top and the confidence interval below. \textsc{GeoMLaMA}, \textsc{GlobalOpinionQA} (\textsc{GOQA}), \textsc{Global-MMLU-CS}, \textsc{NormAD}, and \textsc{BLEnD} entries are percentage-point accuracy gains. \textsc{WorldValuesBench} (\textsc{WVB}) reports Earth Mover Distance reduction, computed as metadata-free EMD minus metadata-formatted EMD, so positive values are better in every column. \textsc{MAPLE} rows compare \tableprimaryplus{$(\Tplus, \Iplus)$}$-$\tableprimaryminus{$(\Tminus, \Iminus)$}; all other rows compare \tableprimaryplus{$\Iplus$}$-$\tableprimaryminus{$\Iminus$}. Dark bold point estimates have intervals that exclude zero, while lighter entries overlap zero.}
\label{tab:external_eval_summary}
\end{table*}

Table~\ref{tab:localnewsqa_model_gain_summary} reports the \textsc{LocalNewsQA} model matrix separately from the external benchmark matrix. It reports \textsc{LocalNewsQA} target-locale accuracy gains and the two paired ambiguous metrics for the completed gold LocalNewsQA runs. The paired metrics remain stringent. Exact pair requires both target and contrast answers to be correct, while margin switch requires the score margin to flip toward the locale-appropriate answer on both sides of the pair.

Table~\ref{tab:external_eval_summary} reports the full external benchmark matrix as metadata gains with confidence intervals. The base \textsc{MAPLE} pretraining comparison has the same direction on all six datasets and both scales. The reference rows are intentionally broader and more mixed. Inference-time metadata alone helps on some model-dataset pairs, especially several \textsc{Global-MMLU-CS} and \textsc{NormAD} rows, but it is not uniformly beneficial across model families. This supports the central \textsc{MAPLE} comparison while showing that inference-time metadata alone is not uniformly sufficient.

\paragraph{Appendix-only WVS projection probe.}

\begin{table*}[t]
\centering
\scriptsize
\setlength{\tabcolsep}{3.5pt}
\begin{tabular}{@{}lcc@{}}
\toprule
Metric & \tableprimaryplus{(\Tplus, \Iplus)} & \tableprimaryminus{(\Tminus, \Iminus)} \\
\midrule
\multicolumn{3}{l}{\textit{Mean distance to the requested country's human WVS map coordinate (lower is better)}} \\
\textsc{MAPLE} 1B & \tableprimaryplus{\textbf{1.416}} & \tableprimaryminus{6.914} \\
\textsc{MAPLE} 3B & \tableprimaryplus{\textbf{2.606}} & \tableprimaryminus{5.794} \\
\bottomrule
\end{tabular}
\caption{Appendix-only country-conditioned World Values Survey projection on the $15$ countries overlapping the \textsc{MAPLE} training set. We compare the primary \textsc{MAPLE} settings under a bundled metadata prompt with soft country framing and short labeled \textsc{WVS} answer options. Top-3 retrieval remains flat at $3/15$ countries, so we report the more diagnostic target-distance metric.}
\label{tab:wvs-country-steering}
\end{table*}

\begin{figure*}[t]
    \centering
    \includegraphics[width=\textwidth]{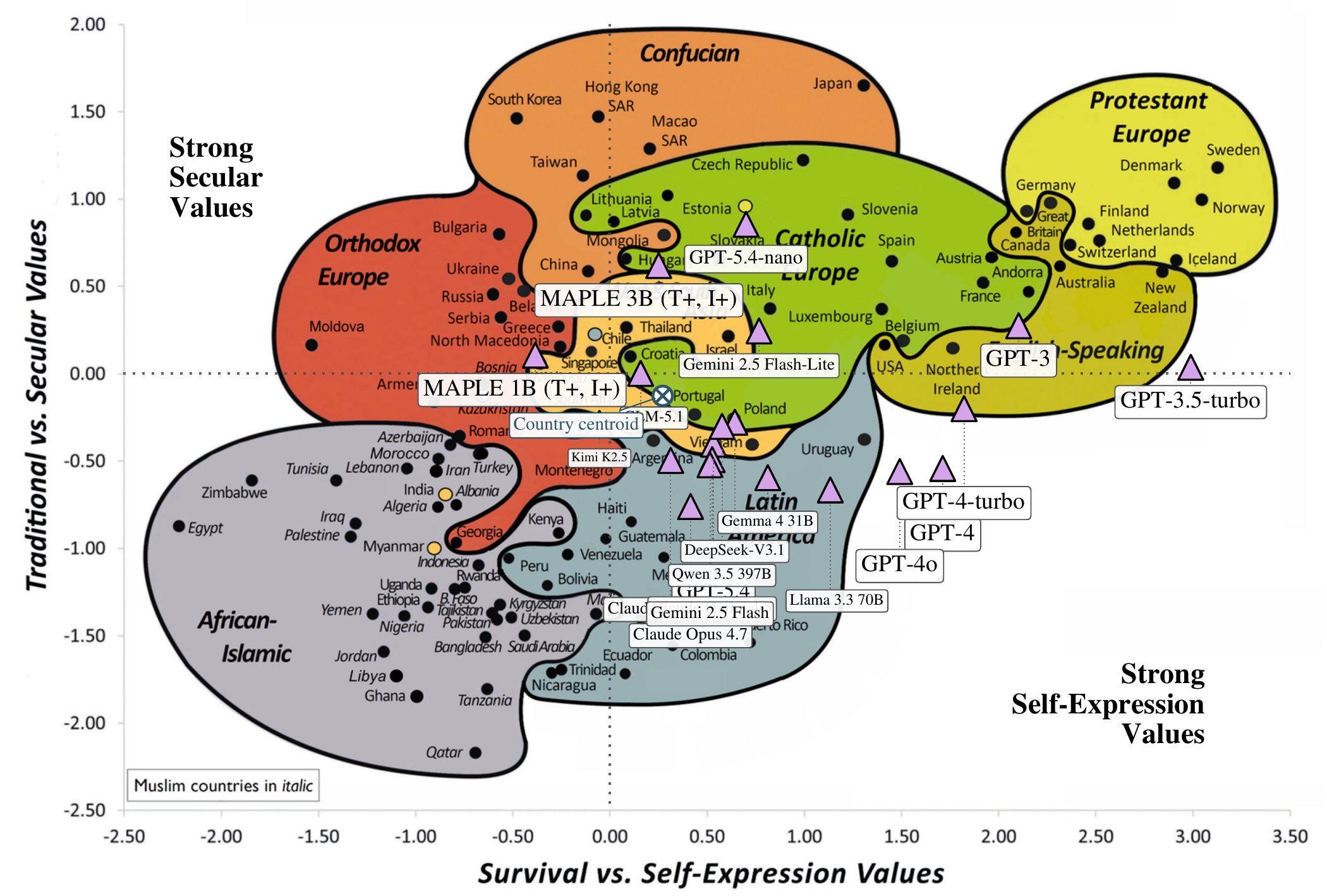}
    \caption{\textbf{Appendix-only all-country \textsc{WVS} map.} Released reference-model centroids are shown alongside the \textsc{MAPLE} centroids averaged over the corresponding country-conditioned sweep. The metadata-conditioned \textsc{MAPLE} centroids fall within the main vendor cloud and remain far closer to the human-country centroid than the metadata-free \textsc{MAPLE} controls, which stay off-map.}
    \label{fig:wvs-country-map}
\end{figure*}

The appendix also reports a \textsc{WVS} projection analysis under a separate country-conditioned protocol. Under this setup, metadata-conditioned \textsc{MAPLE} moves much closer to the requested countries than the metadata-free controls at both scales. Mean target distance falls from $6.91$ to $1.42$ at $1$B and from $5.79$ to $2.61$ at $3$B on the $15$-country overlap set. On the full $107$-country evaluation, the averaged \textsc{MAPLE} $(\Tplus, \Iplus)$ centroids land at $0.695$ and $0.746$ units from the human-country centroid. Those metadata-conditioned centroids are more central than \textsc{Gemma} 4 31B and the older \textsc{GPT} baselines, but they are less central than the strongest recent references in the completed provider evaluation, including \textsc{GLM-5.1}, \textsc{Qwen} 3.5 397B, \textsc{Kimi} K2.5, \textsc{Claude} Sonnet 4.6, and \textsc{GPT-5.4}. The metadata-free \textsc{MAPLE} controls remain off-map. Top-$3$ retrieval remains weak, so we interpret this probe as directional steering toward the requested countries rather than exact country recovery. In that sense, it still provides convergent cross-task evidence that the same metadata-conditioned \textsc{MAPLE} variants that help on \textsc{LocalNewsQA} also induce more target-directed movement on an independent cultural projection task.

\renewcommand{\thesection}{A.\arabic{section}}
\section{Additional Results}
\label{sec:additional-results}

\begin{figure*}[t]
    \centering
    \includegraphics[width=\textwidth]{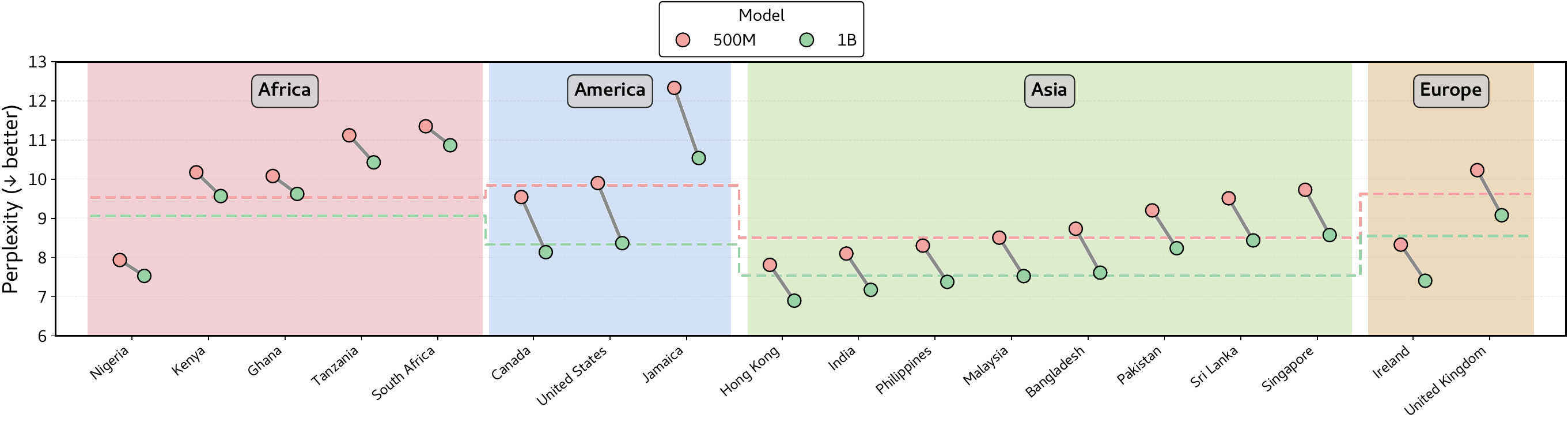}
    \caption{\textbf{Country-level in-region perplexity.} \localwith{} is aggregated within each country by micro-averaged perplexity for the local $500$M and $1$B models.}
    \label{fig:local-country}
\end{figure*}

\begin{figure}[t]
    \centering
    \includegraphics[width=0.48\textwidth]{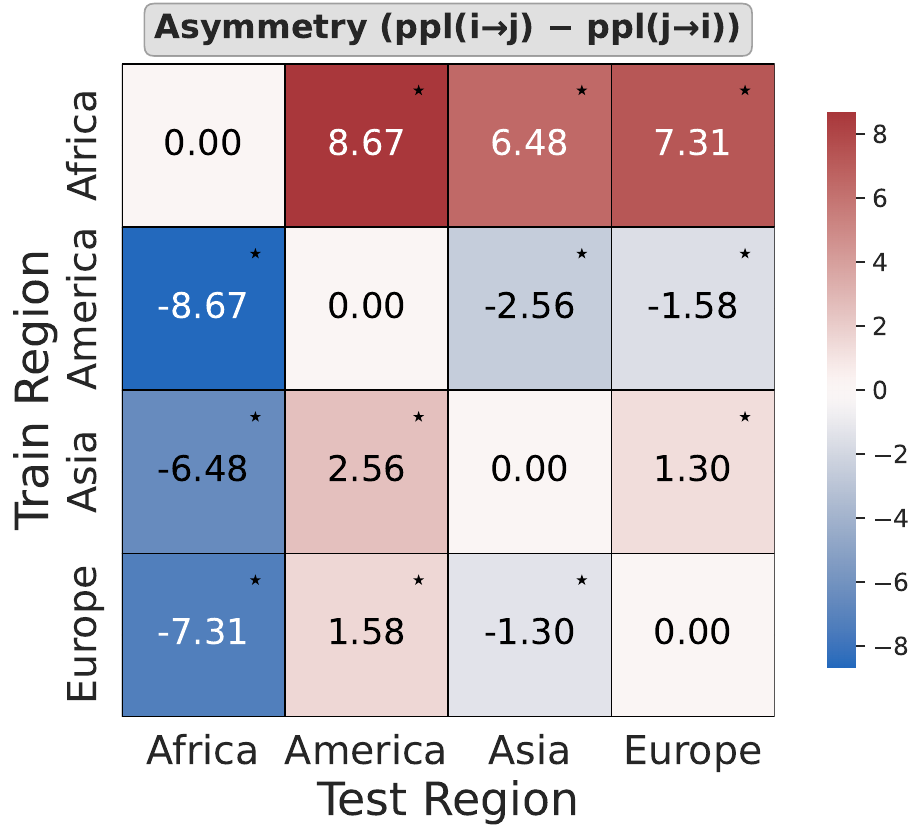}
    \caption{\textbf{Cross-continent asymmetry for $1$B local models.} A higher value indicates that model trained on continent $i$ when evaluated on continent $j$ has a higher perplexity than when a model trained on continent $j$ is evaluated on continent $i$.}
    \label{fig:local-asymmetry}
\end{figure}

\begin{figure}[t]
    \centering
    \includegraphics[width=0.48\textwidth]{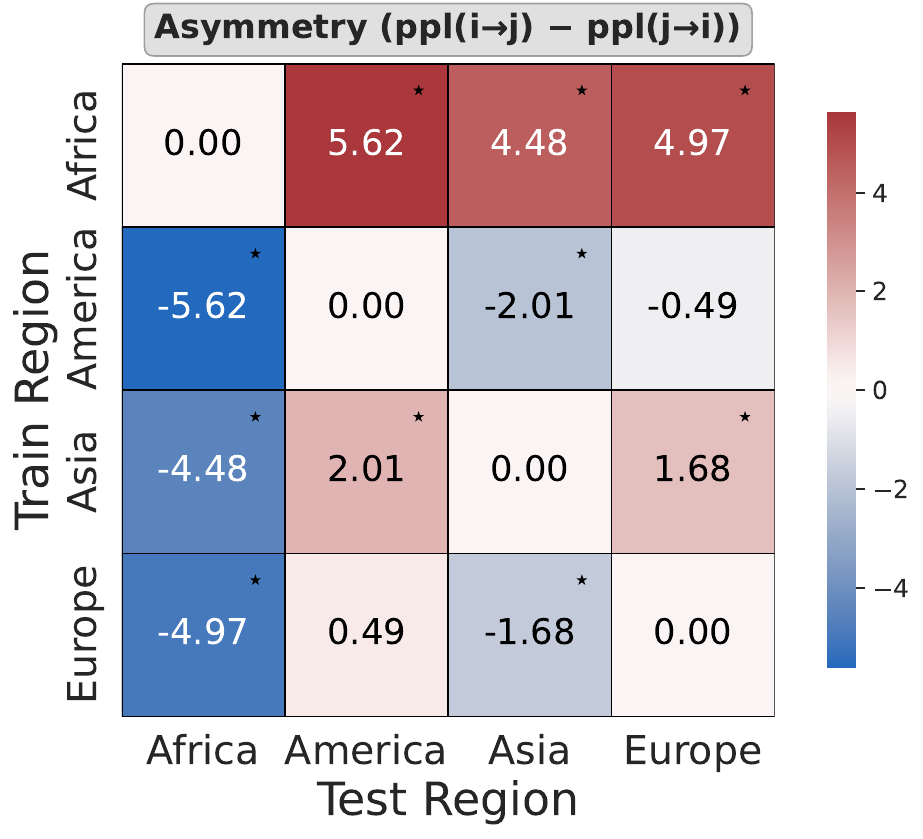}
    \caption{\textbf{Cross-continent asymmetry for $500$M local models.} A higher value indicates that model trained on continent $i$ when evaluated on continent $j$ has a higher perplexity than when a model trained on continent $j$ is evaluated on continent $i$.}
    \label{fig:local-asymmetry-500m}
\end{figure}

\begin{figure}[t]
    \centering
    \includegraphics[width=0.48\textwidth]{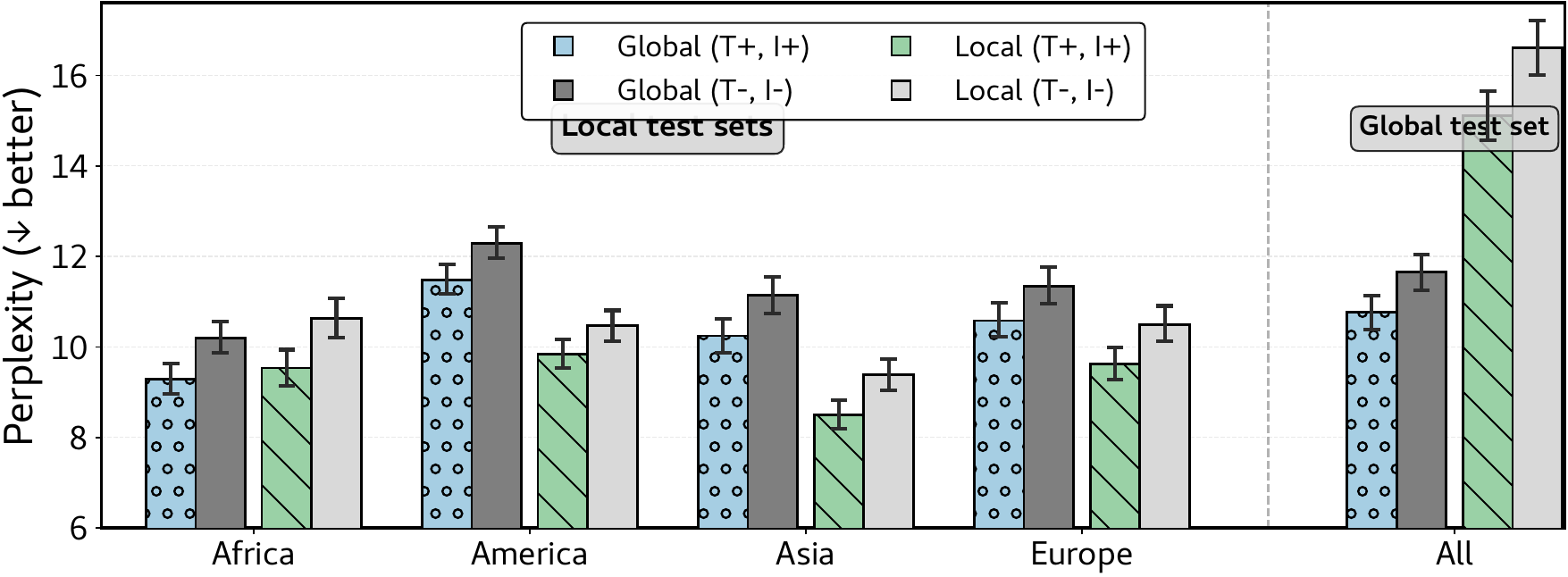}
    \caption{\textbf{Local-vs-global perplexity at $500$M.} \globalwith{} models conditioned with metadata during training and inference have lower perplexities than the control variant \globalwithout{}, and have similar perplexities to the \localwith{} models on local test sets.}
    \label{fig:local-global-500m}
\end{figure}

\begin{figure*}[t]
    \centering
    \begin{subfigure}[t]{0.49\textwidth}
        \centering
        \includegraphics[width=\linewidth]{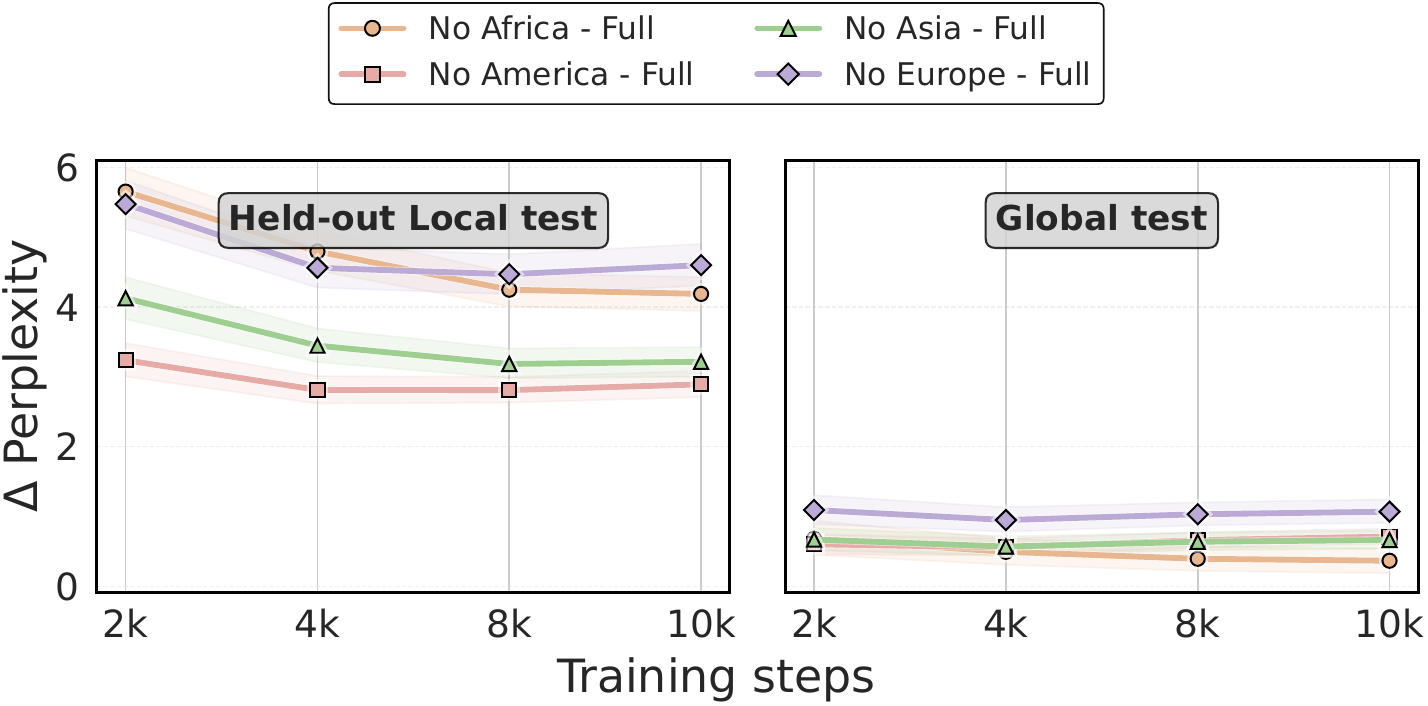}
        \caption{Metadata-conditioned global models.}
    \end{subfigure}
    \hfill
    \begin{subfigure}[t]{0.49\textwidth}
        \centering
        \includegraphics[width=\linewidth]{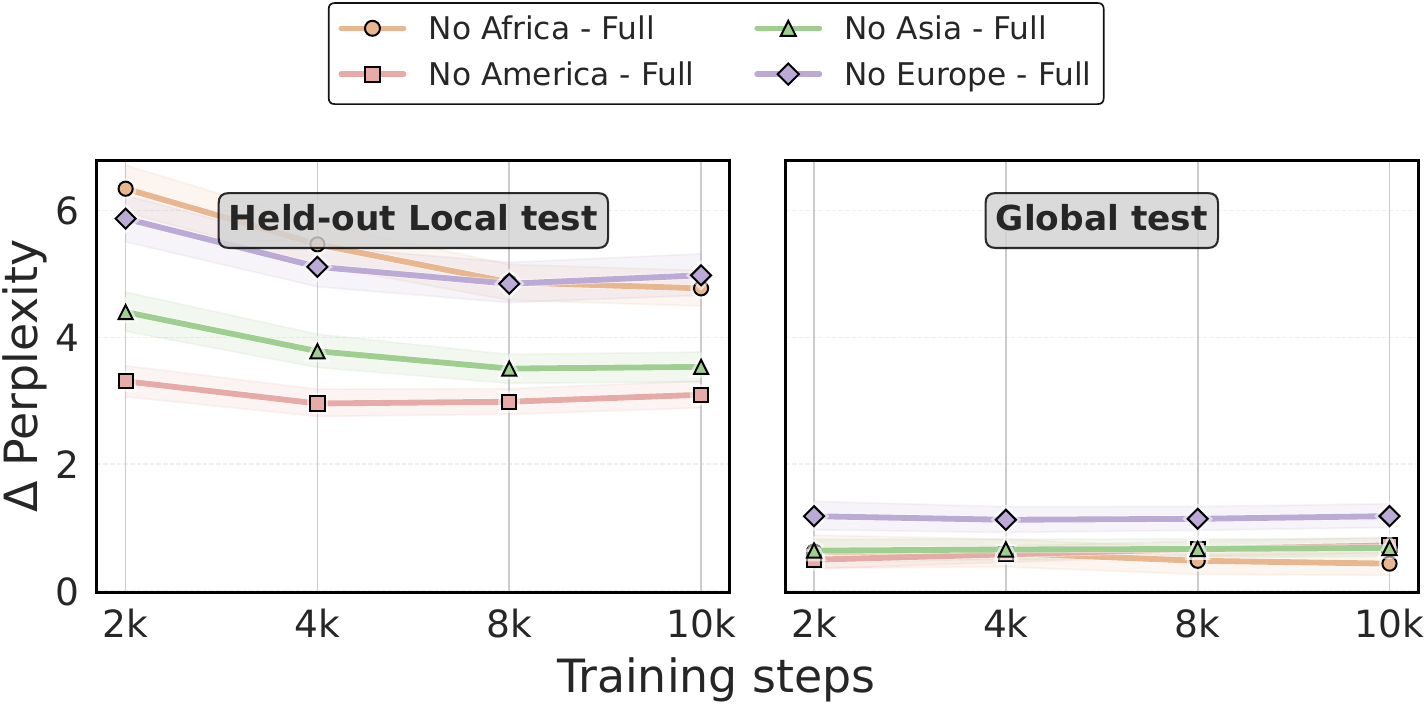}
        \caption{Metadata-free global models.}
    \end{subfigure}
    \caption{\textbf{Leave-one-out global-model ablations.} In both settings, excluding any single continent raises perplexity relative to the corresponding full global model on both the held-out continent and the combined global test set, indicating that metadata formatting does not substitute for geographic coverage. The metadata-conditioned comparison is statistically significant at $95\%$ confidence.}
    \label{fig:ablation-leaveone}
\end{figure*}

\begin{figure*}[t]
    \centering
    \includegraphics[width=\textwidth]{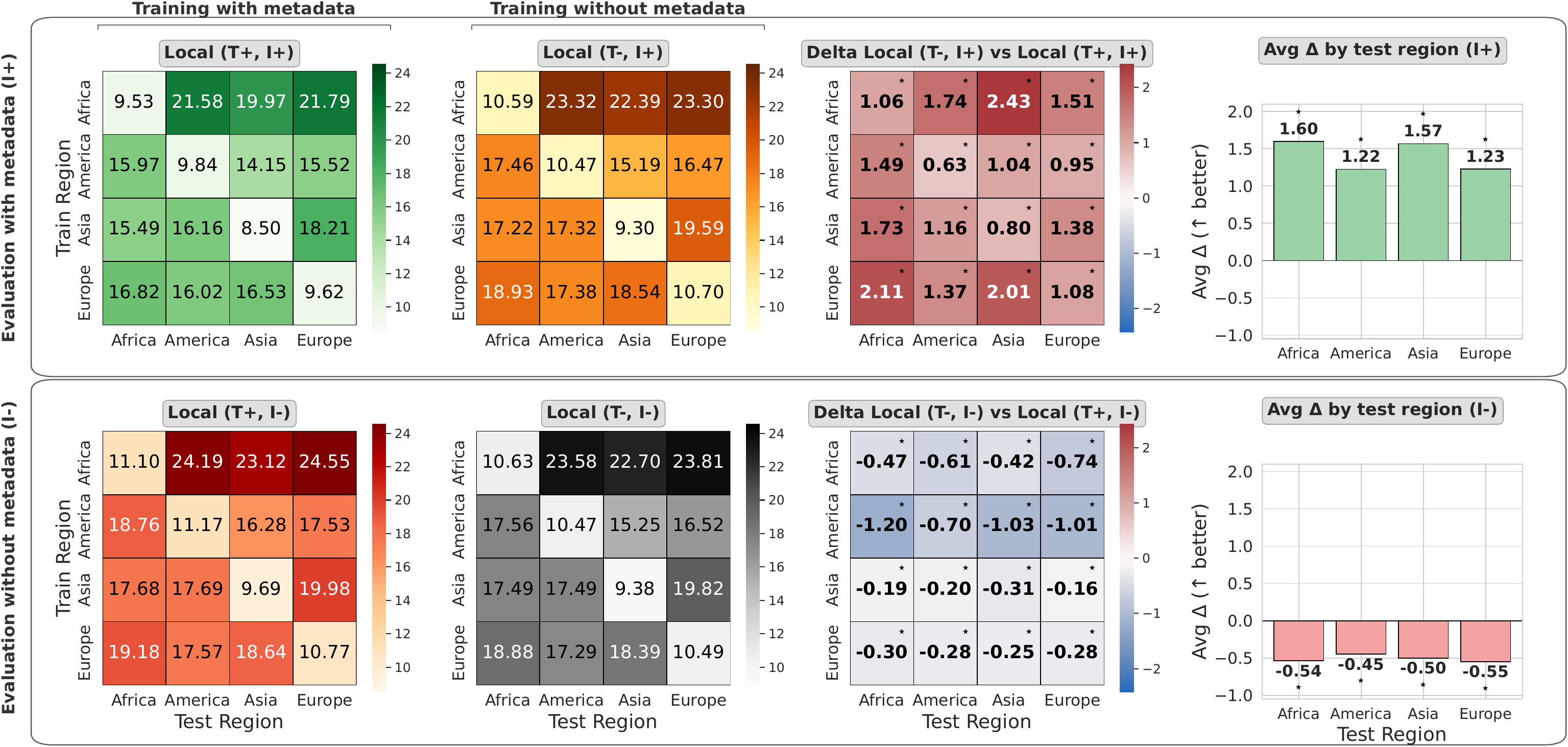}
    \caption{\textbf{Cross-continent perplexity for $500$M local models.} \localwith{} models have higher perplexities on cross-continent test sets (off-diagonal) than on local ones. Positive differences between perplexities of models trained without metadata and with metadata on the same test sets indicate the effectiveness of metadata in improving cross-region generalization while maintaining local performance.}
    \label{fig:local-crosstest-500m}
\end{figure*}

Figures~\ref{fig:local-country}-\ref{fig:local-crosstest-500m} collect the remaining supporting views for the main controlled language-modeling studies. They include country-level local performance, asymmetry across continent pairs, the $500$M counterpart to the local-vs-global comparison, both leave-one-out controls, and the supplementary $500$M cross-continent plot. We place them together here so that each supporting result remains close to the figure that substantiates it.

Balanced $20$k-document samples suggest why the Africa-trained models are least symmetric. Africa documents are shorter on average ($243$ tokens versus $254$ to $341$ elsewhere), have markedly lower cross-continent named-entity overlap with the other regions ($0.149$ to $0.210$ versus $1.000$ within-Africa), and concentrate more heavily in one dominant theme (\textit{Development \& Society} at $0.756$ versus $0.569$ to $0.634$ elsewhere).

\begin{figure*}[t]
    \centering
    \includegraphics[width=\textwidth]{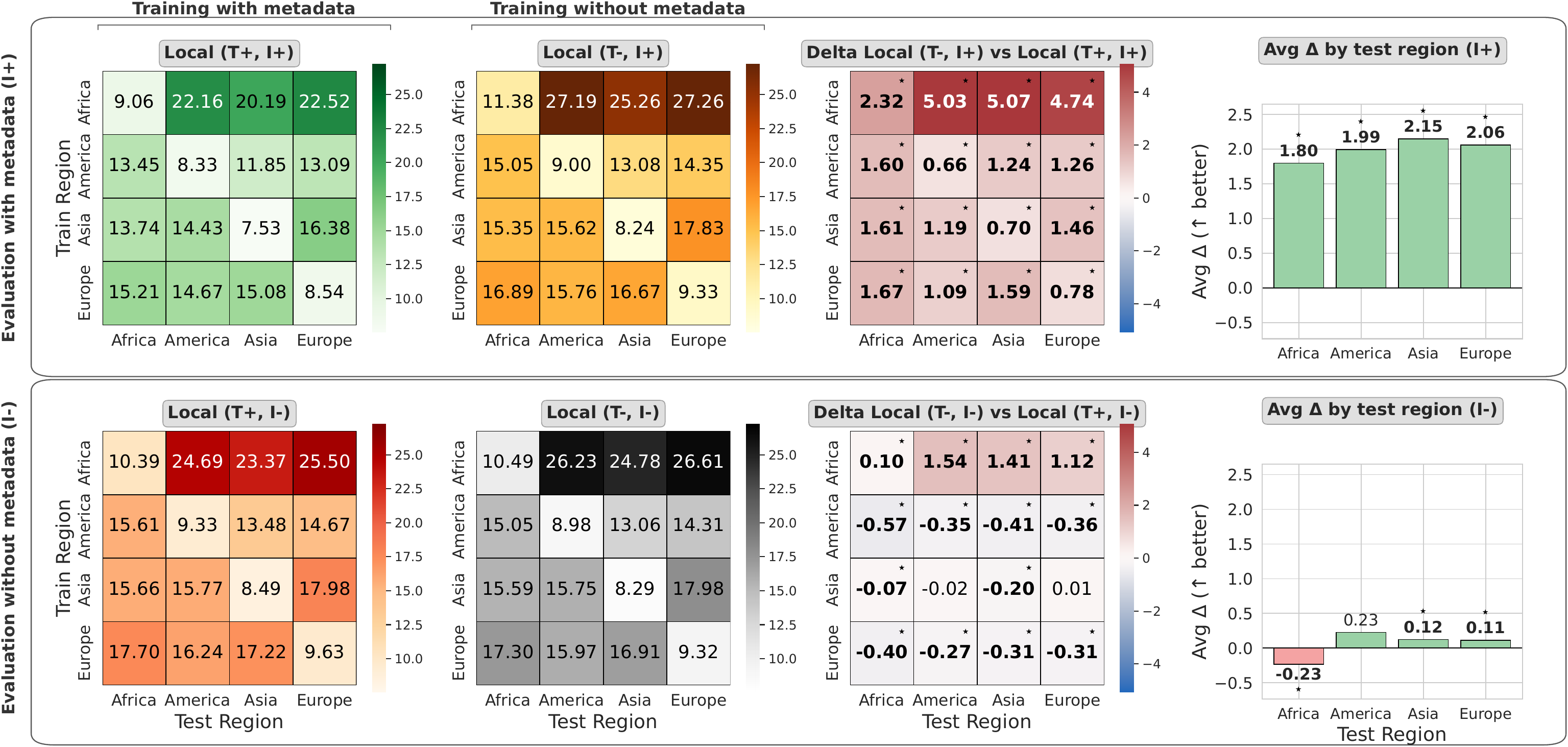}
    \caption{\textbf{Full cross-continent diagnostic view.} The four $1$B local model configurations show raw heatmaps for the \texttt{I+} and \texttt{I-} settings, same-setting deltas, and average delta by test region. Positive $\Delta$ favors metadata conditioning.}
    \label{fig:local-crosstest-full}
\end{figure*}

\begin{figure*}[t]
    \centering
    \includegraphics[width=\textwidth]{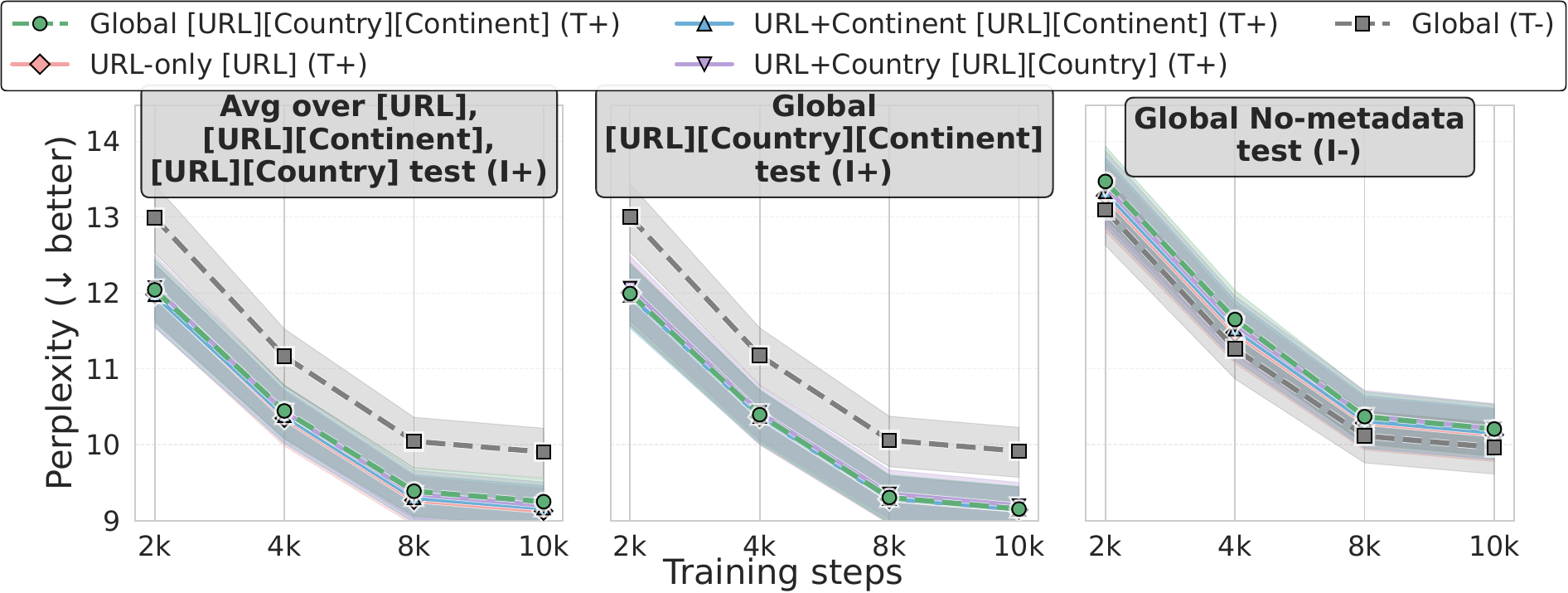}
    \caption{\textbf{Expanded metadata ablations.} Full checkpointed results for \URLChip{}, \URLCountryChip{}, \URLContinentChip{}, \CountryChip{}, and \ContinentChip{}, evaluated on the five metadata-formatted test sets, the \FullMetaChip{} test set, and the no-metadata test set. The same overall pattern persists across intermediate checkpoints, and \URLChip{} remains the most effective single-metadata variant.}
    \label{fig:ablation-meta-all}
\end{figure*}

\paragraph{Scale extension to $3$B, compute, and token efficiency.}

\begin{figure*}[t]
    \centering
    \includegraphics[width=\textwidth]{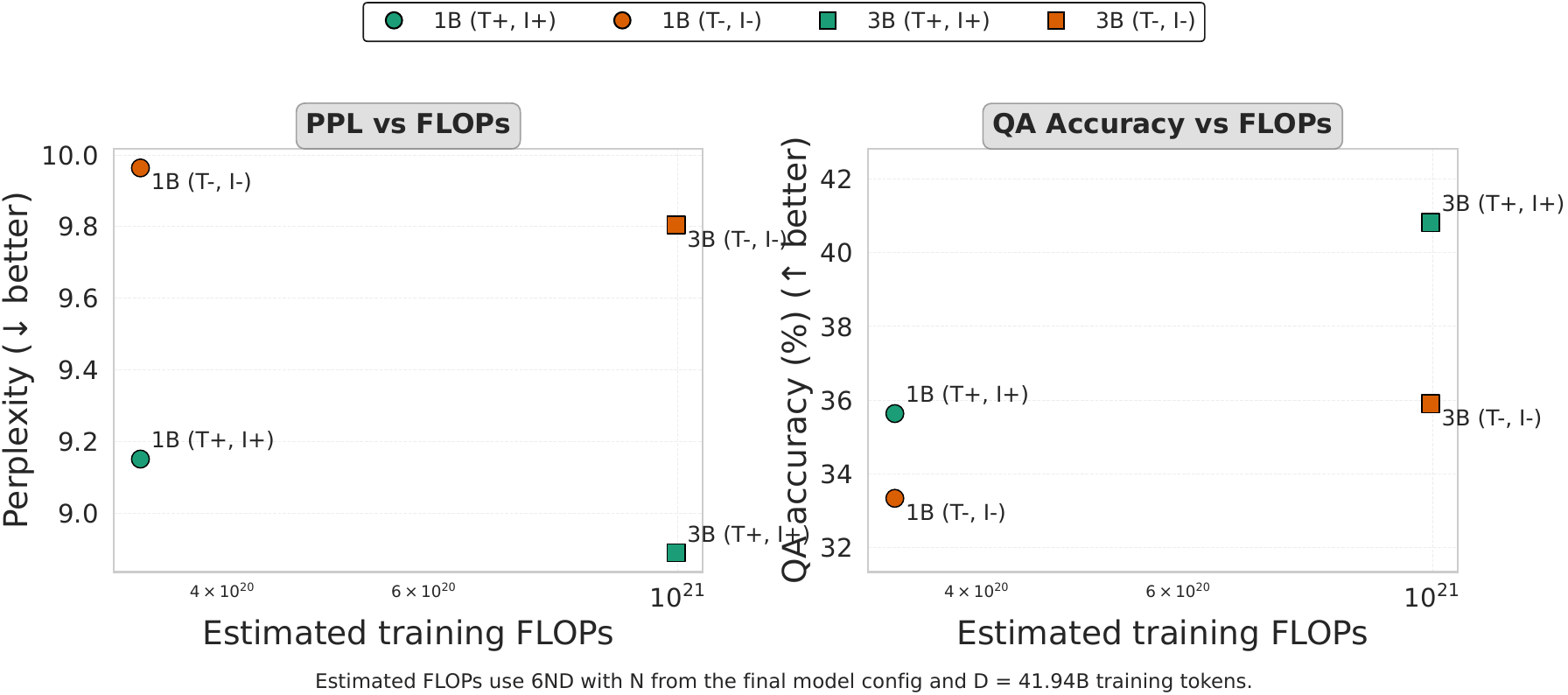}
    \caption{\textbf{Compute/performance tradeoff.} Final performance is plotted against estimated training FLOPs for the global $1$B and $3$B families. The left panel shows global perplexity, and the right panel shows downstream QA accuracy. The $3$B QA points use the same SFT protocol as the $1$B appendix comparison.}
    \label{fig:compute-tradeoff}
\end{figure*}

Figure~\ref{fig:compute-tradeoff} summarizes the final global-model tradeoff between pretraining compute and downstream performance for the $1$B and $3$B families. We plot global perplexity together with downstream QA accuracy, using estimated training FLOPs ($6ND$) because the $3$B runs used a separate $4\times$B200 setup while the $500$M and $1$B families used $4\times$A100s. Absolute runtime differs accordingly. The successful $3$B pretraining runs used a single node with four B200 GPUs, whereas the smaller families used single nodes with four A100 GPUs. The optimizer, schedule, tokenizer, and data pipeline remain the same, so we interpret this hardware difference as relevant to throughput rather than to the factorial comparison itself. For an apples-to-apples comparison, the $3$B QA points use the same SFT protocol as the $1$B appendix comparison. Figure~\ref{fig:token-efficiency} complements this view by showing the training-token trajectories for $500$M, $1$B, and $3$B. Across all three scales, \globalwith{} reaches the final \globalwithout{} perplexity using fewer training tokens, showing that metadata improves optimization efficiency rather than only the final value.

\paragraph{Country-level breakdowns.}

\begin{figure*}[t]
    \centering
    \begin{subfigure}[t]{0.98\textwidth}
        \centering
        \includegraphics[width=\textwidth]{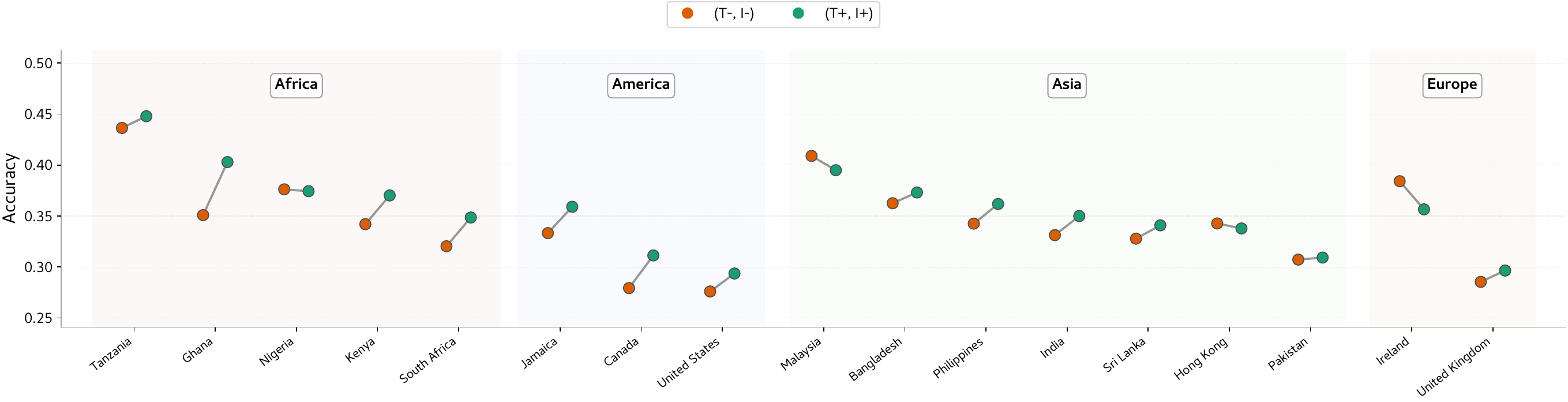}
        \caption{$1$B chat models.}
        \label{fig:country-qa-1b}
    \end{subfigure}
    \vspace{0.2em}
    \begin{subfigure}[t]{0.98\textwidth}
        \centering
        \includegraphics[width=\textwidth]{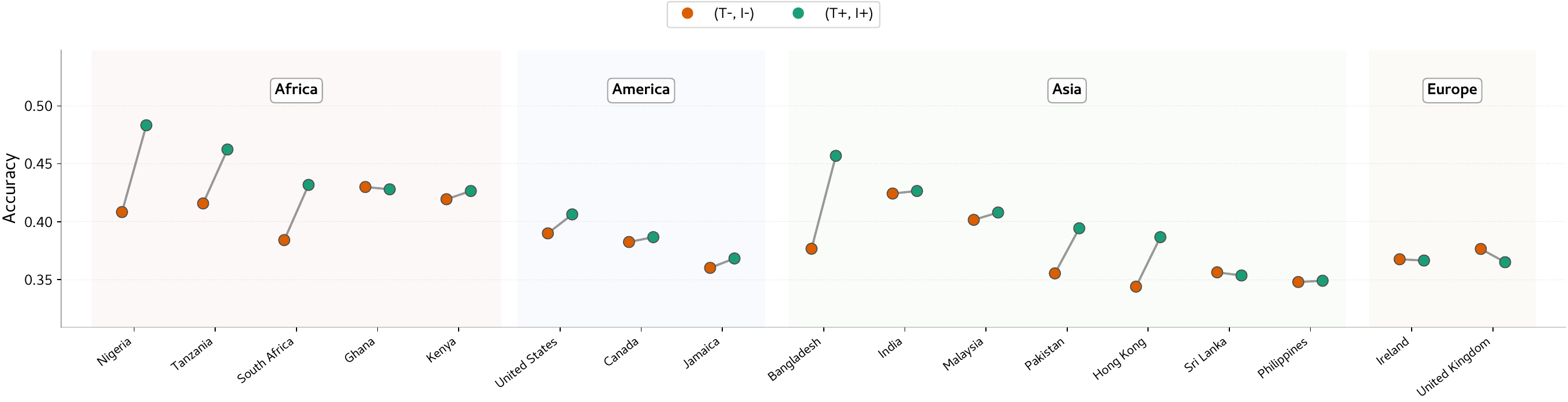}
        \caption{$3$B chat models.}
        \label{fig:country-qa-3b}
    \end{subfigure}
    \caption{\textbf{Country-level downstream QA accuracy.} Global chat models are evaluated after SFT. Within each country, \globalwith{} and \globalwithout{} are compared over the same question pool; the larger model family shows stronger country-to-country variance.}
    \label{fig:country-qa}
\end{figure*}

The local-model analyses in Section~\ref{sec:exp1} operate on continent-level train/test splits, but the \texttt{with\_metadata} test samples preserve the underlying \texttt{COUNTRY} tag in the serialized text. This lets us aggregate performance by country after evaluation. For perplexity, we micro-average by summing token-level losses and token counts within each country before exponentiating. The main paper's Figure~\ref{fig:local-country} includes this country-level view for the local $500$M and $1$B models. Figure~\ref{fig:country-qa} provides the analogous country-level downstream accuracy breakdown for the global chat models after SFT.

\paragraph{Adversarial metadata corruption during downstream inference.}
The main downstream experiment uses correct locale metadata at inference. To test robustness, we repeat the same benchmark while corrupting the URL (and therefore the corresponding country and continent tags) for $25\%$, $50\%$, $75\%$, and $100\%$ of the evaluation samples.

\begin{figure*}[t]
    \centering
    \begin{subfigure}[t]{0.49\textwidth}
        \centering
        \includegraphics[width=\linewidth]{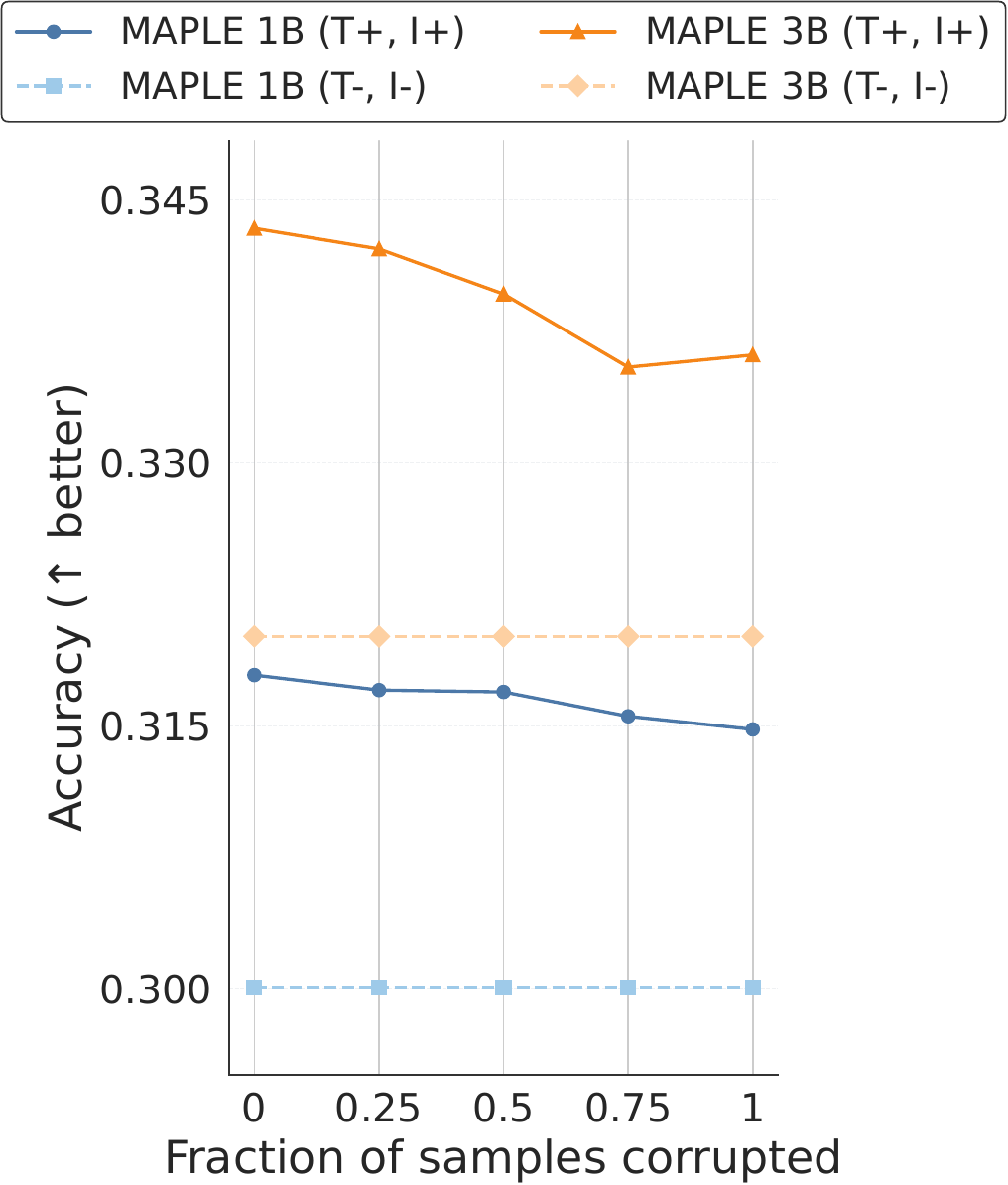}
        \caption{Adversarial URL corruption.}
        \label{fig:adversarial-qa}
    \end{subfigure}
    \hfill
    \begin{subfigure}[t]{0.49\textwidth}
        \centering
        \includegraphics[width=\linewidth]{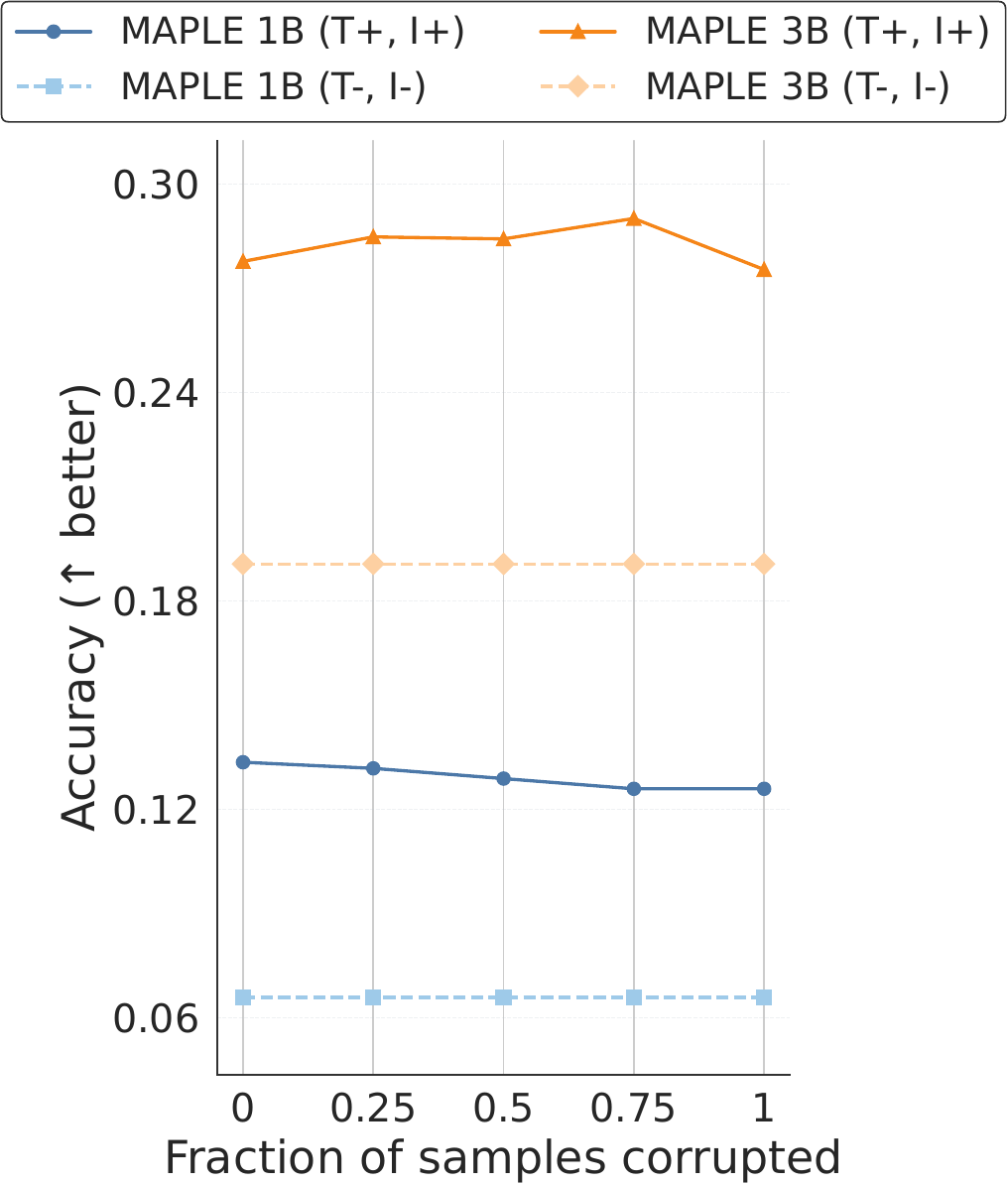}
        \caption{Ambiguous split only.}
        \label{fig:adversarial-qa-noexplicit}
    \end{subfigure}
    \caption{\textbf{Adversarial downstream evaluation.} Pretrained closed-form scoring is evaluated under metadata corruption on the final gold \textsc{LocalNewsQA} split. Figure~\subref{fig:adversarial-qa} reports full-set target accuracy, and Figure~\subref{fig:adversarial-qa-noexplicit} reports the ambiguous split only. Increasing metadata corruption reduces metadata-conditioned inference, while metadata-free inference is largely unchanged. The narrower URL-country-mask variant remains closer to the clean run than to the metadata-free control, indicating that country-code cues matter but broader URL structure also contributes.}
    \label{fig:adversarial-qa-pair}
\end{figure*}

\end{document}